\documentclass[10pt,twocolumn,letterpaper]{article}

\usepackage[pagenumbers]{cvpr} 

\usepackage[dvipsnames,table]{xcolor}

\definecolor{cvprblue}{rgb}{0.21,0.49,0.74}
\usepackage[pagebackref,breaklinks,colorlinks,citecolor=cvprblue]{hyperref}

\usepackage{fdsymbol}
\usepackage{multirow}
\usepackage{bbm}
\usepackage{arydshln}

\usepackage{subcaption}

\DeclareMathAlphabet\mathbfcal{OMS}{cmsy}{b}{n}

\definecolor{cgreen}{HTML}{39b54a}  
\definecolor{cyellow}{HTML}{FFC000}
\definecolor{cred}{HTML}{A10035}
\definecolor{cpurple}{HTML}{7030A0}
\definecolor{aliceblue}{rgb}{0.94, 0.97, 1.0}

\usepackage{pifont}
\newcommand{\cmark}{\textcolor{purple}{\ding{52}}}%

\definecolor{voc_cow}{HTML}{0C1E7F}
\definecolor{voc_horse}{HTML}{FB2576}
\definecolor{violet}{HTML}{BB1AEF}

\usepackage{makecell}

\newcommand{\pub}[1]{\color{gray}{\tiny{#1}}}

\usepackage{tabulary}
\newcolumntype{I}{!{\vrule width 1pt}}

\newcolumntype{x}[1]{>{\centering\arraybackslash}p{#1pt}}
\newcolumntype{y}[1]{>{\raggedright\arraybackslash}p{#1pt}}
\newcolumntype{z}[1]{>{\raggedleft\arraybackslash}p{#1pt}}
\newlength\savewidth

\makeatletter
\newcommand{\thickhline}{%
	\noalign {\ifnum 0=`}\fi \hrule height 1pt
	\futurelet \reserved@a \@xhline
}
\makeatother

\newcommand{\increase}[1]{
	{\fontsize{6pt}{0.5em}\selectfont\color{purple}{$\uparrow$~{#1}}}
}
\newcommand{\decrease}[1]{
	{\fontsize{6pt}{0.5em}\selectfont\color{gray!48}{$\downarrow$~{#1}}}
}

\newcolumntype{x}[1]{>{\centering\arraybackslash}p{#1pt}}
\newcolumntype{y}[1]{>{\raggedright\arraybackslash}p{#1pt}}
\newcolumntype{z}[1]{>{\raggedleft\arraybackslash}p{#1pt}}

\newcommand{\bmhead}[1]{\noindent\textbf{#1}}

\usepackage[misc]{ifsym}

\usepackage{fontawesome}

\title{Instance Brownian Bridge as Texts for Open-vocabulary Video Instance Segmentation}

\author{
Zesen Cheng$^{1}$\thanks{\Letter\ Corresponding Author\ \ \ \faGithub\ \href{https://github.com/sennnnn/OpenVIS}{github.com/sennnnn/OpenVIS}} \quad Kehan Li$^{1}$ \quad Hao Li$^{1}$ \quad Peng Jin$^{1}$ \\
Chang Liu$^{3}$ \quad Xiawu Zheng$^{4}$ \quad Rongrong Ji$^{4}$ \quad Jie Chen$^{1,2}$~\textsuperscript{\Letter} \and
$^{1}$ School of Electronic and Computer Engineering, Peking University \\
$^{2}$ Peng Cheng Laboratory \quad
$^{3}$ Tsinghua University \quad 
$^{4}$ Xiamen University \\
}
\renewcommand\footnotemark{}

\begin{document}
\maketitle
\begin{abstract}
Temporally locating objects with arbitrary class texts is the primary pursuit of open-vocabulary Video Instance Segmentation~(VIS).
Because of the insufficient vocabulary of video data, previous methods leverage image-text pretraining model for recognizing object instances by separately aligning each frame and class texts, ignoring the correlation between frames.
As a result, the separation breaks the instance movement context of videos, causing inferior alignment between video and text.
To tackle this issue, we propose to link frame-level instance representations as a Brownian \underline{\textbf{Bri}}dge to model instance dynamics and align bridge-level instance representation to class texts for more precisely open-vocabulary \underline{\textbf{VIS}}~(\textbf{BriVIS}).
Specifically, we build our system upon a frozen video segmentor to generate frame-level instance queries, and design \underline{\textbf{T}}emporal \underline{\textbf{I}}nstance \underline{\textbf{R}}esampler~(\textbf{TIR}) to generate queries with temporal context from frame queries.
To mold instance queries to follow Brownian bridge and accomplish alignment with class texts, we design \underline{\textbf{B}}ridge-\underline{\textbf{T}}ext \underline{\textbf{A}}lignment~(\textbf{BTA}) to learn discriminative bridge-level representations of instances via contrastive objectives. 
Setting MinVIS as the basic video segmentor, BriVIS surpasses the Open-vocabulary SOTA~(OV2Seg) by a clear margin. For example, on the challenging large-vocabulary VIS dataset~(BURST), BriVIS achieves 7.43 mAP and exhibits 49.49\% improvement compared to OV2Seg~(4.97 mAP).
\end{abstract}
\begin{figure}[t]
\centering
\begin{subfigure}[b]{0.48\textwidth}
    \centering
    \includegraphics[width=1.0\textwidth]{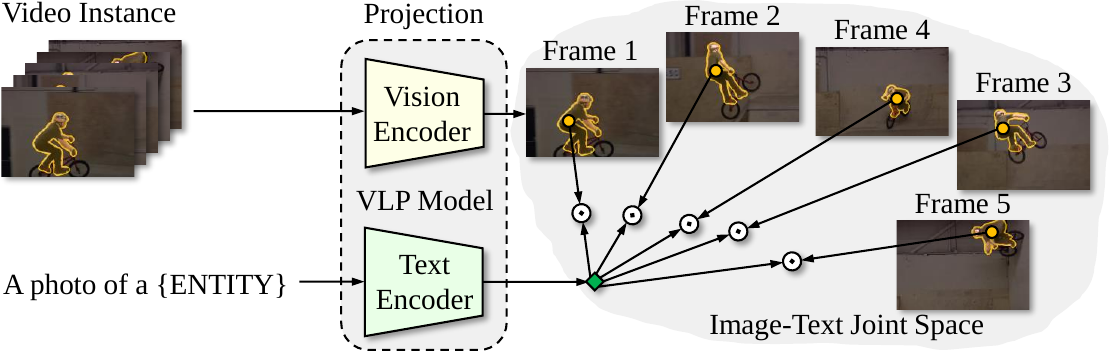}
    \caption{Frame-text Alignment}
    \label{fig:motivation_fram}
\end{subfigure}
\par\medskip
\begin{subfigure}[b]{0.48\textwidth}
    \centering
    \includegraphics[width=1.0\textwidth]{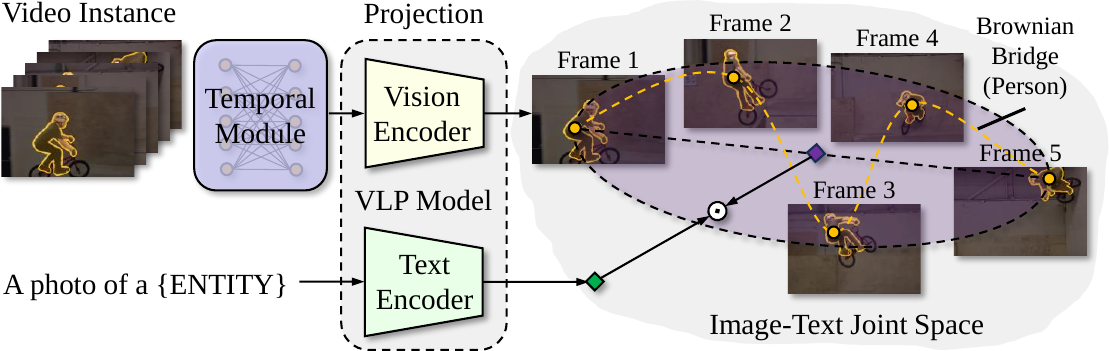}
    \caption{Bridge-text Alignment}
    \label{fig:motivation_brid}
\end{subfigure}
\caption{\textbf{The mechanical difference} between (a) frame-text and (b) bridge-text~(ours) alignment. 
Because of the deficient vocabulary of video data, OVVIS adopts image-text VLP model to provide semantic space.
Previous methods recognize instances by integrating frame-text alignment results.
Our method links frame-level instance features as a Brownian bridge and aligns the bridge center to class texts to consider instance movement information when recognizing video instances.
\textbf{\color{cyellow}{Yellow circles}}, \textbf{\color{cgreen}{Green circles}}, and \textbf{\color{cpurple}{Diamond}} denote frame-level instance features, class text features, and bridge center. $\odot$ denotes calculating alignment score.
}
\vspace{-10pt}
\label{fig:motivation}
\end{figure}
\section{Introduction}

Video Instance Segmentation~(VIS) aims at classifying, segmenting, and tracking all object instances in the input videos~\cite{yang2019video}.
Based on the VIS task, the Open-Vocabulary Video Instance Segmentation~(OVVIS)~\cite{wang2023towards,guo2023openvis} aims to recognize any object categories for adapting real-world application scenarios.
The development of image-text Vision-Language Pretraining~(VLP) models~\cite{jia2021scaling,yu2022coca,yuan2021florence,li2022blip,li2022grounded,sun2023eva}~(e.g., CLIP~\cite{radford2021learning}) largely boosts image-level open-vocabulary tasks~\cite{gu2021open,huynh2022open,xu2022simple}, which attracts researchers to adapt image-text VLP models for open-vocabulary video tasks.
Albeit directly constructing video-text VLP models can also significantly advance open-vocabulary video tasks, it is not economical due to the expensiveness of large-scale video-text pairs collection and annotation.
Therefore, previous OVVIS models mainly use image-text VLP models for achieving open-vocabulary recognition.

To fit the image-text pretraining models' input modality, initial efforts~\cite{wang2023towards,guo2023openvis} of OVVIS propose to depart video into frames for leveraging the image-text VLP to recognize video instances via ensembling the instance-text alignment score of each frame~(Fig.~\ref{fig:motivation_fram}).
Nevertheless, this route suffers from a major issue.
Lacking the temporal modeling ability between frames, VLP models ignore how instance features evolve over time when aligning class text and instance features in a single frame.
This results in suboptimal video instance recognition because the spatial-temporal context information contained in the video dynamics has been broadly demonstrated to be significant for instance semantic description~\cite{tu2017video,hou2021bicnet,ding2022language,hui2021collaborative,hui2023language}.

To remedy this, we first assume the movement of instance as a Brownian motion because instance features of neighbor frames share highly similarity~\cite{huang2022minvis}.
Then we find that the instance movement contains a strong causal dependency during the process, i.e., the middle instance state can be easily inferred by observing the change between the start and end instance states, which agrees with the goal-conditioned property of Brownian bridge and support us to further model the Brownian motion as a Brownian bridge~\cite{revuz2013continuous,wang2022language,wang-etal-2023-dialogue,zhang2023modeling}.
According to the goal-conditioned property, the average of the start and end instance features, i.e., the bridge center, is able to roughly represent the whole instance movement. 
Subsequently, we align the bridge center to class texts, which enables considering instance dynamics in image-text space for recognizing video instances. 
The overall mechanism of this \underline{\textbf{B}}ridge-\underline{\textbf{T}}ext \underline{\textbf{A}}lignment~(\textbf{BTA}) is illustrated in Fig.~\ref{fig:motivation_brid}.

Specifically, the calculation flow of BTA contains three steps.
Firstly, we constrain the Brownian bridge width by adopting hinge loss to modulate the distance between the head and tail instance features.
Secondly, we derive a bridge-based contrastive objective to pull the middle frame features to follow the bridge distribution and push the middle frame features of other instances away.
Finally, we adopt a contrastive objective to guide bridge center close to the corresponding class text and away from irrelevant class texts.
Furthermore, since BTA serves as a training objective, training the whole model with multiple frames input requires a heavy cost.
To reduce the cost, we propose to freeze a pretrained VIS segmentation model and build a \underline{\textbf{T}}emporal \underline{\textbf{I}}nstance \underline{\textbf{R}}esampler~(\textbf{TIR}) upon on the segmentor to learn how to capture instance dynamics.
Associating BTA and TIR with a frozen pretrained VIS method, we build an effective OVVIS system, which is termed \textbf{BriVIS}.

To check the effectiveness of our method, we set MinVIS~\cite{huang2022minvis} as our basic segmentor and conduct extensive experiments.
In our experiments, we mainly set a large-vocabulary VIS dataset~(LV-VIS~\cite{wang2023towards}) as our training set to learn vocabulary generalization ability and adopt “zero-shot” manner to evaluate the open-vocabulary performance on regular VIS datasets~(Youtube-VIS 2019~\cite{yang2019video}, Youtube-VIS 2021~\cite{yang2019video}, BURST~\cite{athar2023burst}, and OVIS~\cite{qi2022occluded}).
As a result, our method surpasses previous OVVIS methods with a clear margin.
For example, our method achieves {7.43} mAP on the challenging large-vocabulary BURST dataset and is +{2.46} mAP better than previous state-of-the-art OV2Seg~\cite{wang2023towards}.
Moreover, our method provides competitive VIS performance when compared to advanced close-vocabulary VIS methods.
These two comparisons fully demonstrate the eminent vocabulary generalization ability of our method. 

In summary, the main contributions are as follows:
\begin{itemize}
\item We propose BTA to model the video instance as a Brownian bridge and then align the bridge center to class text. Because of the goal-conditioned property of the Brownian bridge, BTA enables recognizing instances in image-text space while considering instance dynamics.
\item The learning of BTA requires a sufficient sampling frame number. Therefore, we propose to freeze a pretrained VIS method and design TIR upon this VIS model as a lightweight parameterized model to learn linking frame-level instance features as Brownian bridge.
\item Compared to previous OVVIS methods, ours improves them by a clear margin. For example, our method gets 7.43 mAP on the challenging BURST dataset, which improves previous SOTA~(OV2Seg, 4.97 mAP) by 49.49\%.
\end{itemize}

\begin{figure*}[t]
\centering
\includegraphics[width=1.0\linewidth]{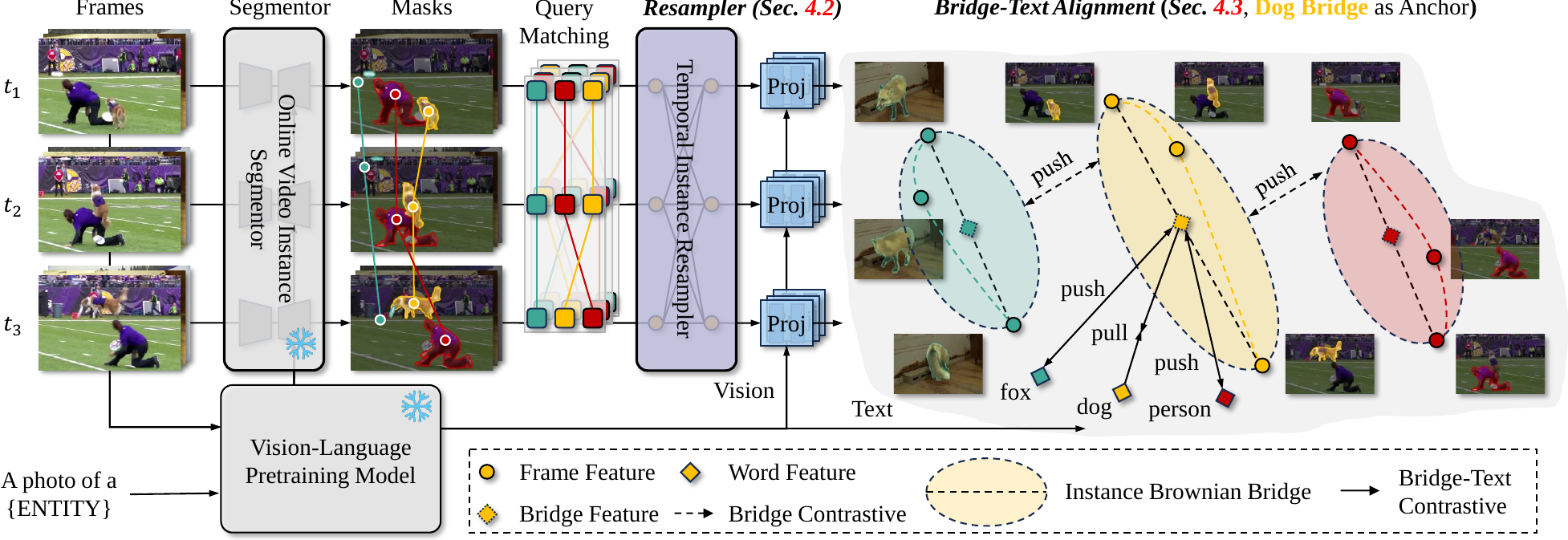} 
\caption{\textbf{The overall pipeline} of our BriVIS. Our main designs are TIR~(Sec.\ref{sec:tir}) and BTA~(Sec.~\ref{sec:bta}). The former regenerates instance queries by building connections between independent instance queries. The latter is used to link instances spanning different frames as a Brownian bridge and align them and text at bridge granularity. The learning of BTA requires sufficient sampling frames, causing expensive computation costs. Therefore, we split training into two stages to pretrain the video segmentor and train TIR and BTA.
}
\label{fig:overall_pipeline}
\vspace{-10pt}
\end{figure*}
\section{Related Works}

\subsection{Video Instance Segmentation}
Video instance segmentation is a comprehensive video recognition task, aiming at segmenting, tracking, and classifying all objects in videos~\cite{yang2019video}.
Transformer-based~\cite{dosovitskiy2020image,carion2020end} VIS methods gradually become the mainstream route because the attention architecture significantly improves the performance and the query-oriented design provides an elegant instance representation.
VisTR~\cite{wang2021end} designs an end-to-end baseline to track the identical object across video via a simple instance query, which is the first attempt to extend DETR~\cite{carion2020end} to the VIS task.
Follow-up works develop in two directions: online and offline.
The online route departs video into frames or clips to process and focuses on how to link objects spanning the adjacent video frames or clips~\cite{huang2022minvis,wu2022defense,zhan2022robust,wu2022efficient,heo2023generalized,zhang2023dvis}. For example, MinVIS associates objects by using Hungarian algorithm~\cite{kuhn1955hungarian} to match queries between two adjacent frames.
The offline route sets whole video as a spatial-temporal volume to process and focuses on designing efficient and effective temporal interaction~\cite{cheng2021mask2former,yang2022temporally,ke2022video,heo2022vita,li2023mdqe}, e.g., IFC~\cite{hwang2021video} designs memory tokens and Seqformer~\cite{wu2022seqformer} proposes frame query decomposition.
Although the offline route achieves more notable performance, the online route is more efficient, especially for long videos.
Therefore, MinVIS~(a simple online transformer-based method) is selected as our segmentor.

\subsection{Open-Vocabulary Segmentation}
Open-vocabulary image segmentation explores how to recognize any categories at pixel level.
Earlier works~\cite{zhao2017open,xian2019semantic,bucher2019zero} attempt to build a pixel-text alignment space by learning to align pixel embedding to word embeddings~\cite{mikolov2013distributed,miller1995wordnet} of class texts.
With the rise of VLP models~\cite{jia2021scaling, yu2022coca, yuan2021florence, li2022blip, li2022grounded, sun2023eva} represented by CLIP~\cite{radford2021learning}, recent research focus shifts to explore how to adapt their superior image-level open-vocabulary ability to pixel-level.
Part of works~\cite{ghiasi2022scaling,li2022languagedriven,liang2023open,zhou2022extract} proposes to directly finetune or distill VLP for remedying this granularity gap, which requires vast segmentation data. 
Another part of works~\cite{ding2023maskclip,han2023open} explores to reuse the original image-text alignment space of VLP. For example, SimpleBaseline~\cite{xu2022simple} proposes to recognize image crops by a frozen CLIP, and SideAdapter~\cite{xu2023side} predicts attention biases to modulate self-attention of CLIP for mask-guided image-text alignment.  
In this paper, we mainly refer to the design philosophy of the latter part of open-vocabulary segmentation works and explore how to adapt VLP for spatial-temporal segmentation tasks.

\subsection{Brownian Bridge Modeling}
Brownian bridge~\cite{revuz2013continuous} is a continuous-time Gaussian stochastic process $B(t)$ whose probability distribution of each time step follows a Gaussian distribution and is conditioned by start state $z_0$ at $t = 0$ and end state $z_T$ at $t = T$:
\begin{equation}
B(t) = \mathcal{N}((1 - \frac{t}{T})z_0 + \frac{t}{T}z_T, \frac{t(T-t)}{T}),
\label{eq:brownian_density}
\end{equation}
where $z_t, t\in[0, T]$ is the middle state of the bridge.
According to Eq.~\ref{eq:brownian_density}, we can find that $z_t$ is approximately the noisy linear interpolation of $z_0$ and $z_T$ modulated by time variable. 
The uncertainty gradually decreases to the lowest at the start side or end side of the bridge and increases to the highest at the bridge center point.

Brownian bridge is a promising tool for modeling process-oriented problems. 
TC~\cite{wang2022language} proposed to generate text in a latent space with Brownian bridge dynamics. 
In this way, the middle text can follow local coherence and be modulated by start and end context, which is the pilot work to leverage the goal-conditioned nature of the Brownian bridge.
Subsequently, the Brownian bridge is further explored in fine-grained video self-supervised learning~\cite{zhang2023modeling} and dialogue generation~\cite{wang-etal-2023-dialogue}.
In our work, we adopt the Brownian bridge to model instance dynamics based on frame-level instance features, which improves the alignment between the instance and the class text in the image-text semantic space when recognizing video instances.

\section{Preliminaries}
Given a video with $T$ frames $\mathbfcal{V} = \{I_1\in\mathbb{R}^{H\times W\times 3}, I_2, ..., I_T\}$ and arbitrary class text set $\mathbfcal{C}$, OVVIS is expected to segment, track, and classify all video instances, which is almost simultaneously introduced in OpenVIS~\cite{guo2023openvis} and OV2Seg~\cite{wang2023towards}.
A video instance represents an identical object spanning valid frames and is determined by two factors: temporal mask $\mathbfcal{M}$ for locating an object at the spatial-temporal level and category $c\in\mathbfcal{C}$.
During the training phase, OVVIS model is provided with finite vocabulary to learn the vocabulary transfer ability.
During the inference phase, OVVIS model is used to recognize the objects of videos with arbitrary categories, even if these categories never appear in the training vocabulary.

\section{Methods}\label{sec:method}

We term our method as BriVIS.
We first describe the overall pipeline of BriVIS in Sec.~\ref{sec:overall_pipe} and then describe the key components: TIR in Sec.~\ref{sec:tir} and BTA in Sec.~\ref{sec:bta} 

\subsection{Overall Pipeline}
\label{sec:overall_pipe}

To provide a concrete and vivid description of our method, we illustrate our overall pipeline in Fig.~\ref{fig:overall_pipeline}.

\bmhead{Online Video Instance Segmentor.} 
We choose MinVIS~\cite{huang2022minvis} as our basic segmentor to generate masks and queries for each frame, whose architecture is essentially an instance segmentation network~(i.e., Mask2Former~\cite{cheng2021mask2former}) with a tracking module~(i.e., Hungarian Query Matching).
Mask2former is comprised of three parts: image backbone, pixel decoder, and query decoder. The video is first departed into 
frames for extracting backbone features $\{\mathbfcal{F}^{b_1}_t\in\mathbb{R}^{\frac{H}{8}\times \frac{W}{8}\times d_1}, \mathbfcal{F}^{b_2}_t\in\mathbb{R}^{\frac{H}{16}\times \frac{W}{16}\times d_2}, \mathbfcal{F}^{b_3}_t\in\mathbb{R}^{\frac{H}{32}\times \frac{W}{32}\times d_3}\}$, where $H$, $W$, and $d$ denote height, width, and feature dimension, respectively. 
Then a pixel decoder gradually integrates backbone features to generate high-quality multi-scale pixel features. 
$3,6,9$-th layer features of CLIP model, meanwhile, are injected in three stages of pixel decoder to enhance the open-vocabulary ability:
\begin{equation}
\mathbfcal{F}^{o_i}_{t} = \texttt{MSDefAttn}(\texttt{MLP}(\mathbfcal{F}^{b_i}_{t}) + \texttt{MLP}(\mathbfcal{F}^{c_i}_{t})),
\end{equation}
where $\texttt{MSDefAttn}$ denotes multi-scale deformable transformer, $\mathbfcal{F}^{o_i}_t$, $\mathbfcal{F}^{b_i}_{t}$, and $\mathbfcal{F}^{c_i}_{t}$ respectively denote the $i$-th stage open-vocabulary, backbone, and CLIP features of $t$-th frame. 
The shape of $\mathbfcal{F}^{c_i}_{t}$ is processed to align $\mathbfcal{F}^{b_i}_{t}$. The spatial shape of $\mathbfcal{F}^{o_i}_{t}$ is equal to $\mathbfcal{F}^{b_i}_{t}$, but its dimension is processed to $d$, e.g., $\mathbfcal{F}^{o_1}_{t}\in\mathbb{R}^{\frac{H}{8}\times \frac{W}{8}\times d}$.
Moreover, pixel decoder also generates pixel embeddings $\mathbfcal{E}^p_t$ of each frame.
Subsequently, $\mathbfcal{F}^{o_i}_t$ is used to attend cross attention of each transformer decoder layer in query decoder for acquiring instance queries $\mathbfcal{Q}_t \in\mathbb{R}^{N \times d}$ and masks $\mathbfcal{M}_t\in\{0, 1\}^{N \times H \times W}$, where $N$ denotes the number of instances.
Finally, we match queries between adjacent frames via Hungarian algorithm~\cite{kuhn1955hungarian} for acquiring resorted video instance queries $\mathbfcal{Q}\in\mathbb{R}^{N \times T \times d}$ and masks $\mathbfcal{M}\in\{0, 1\}^{N \times T \times H \times W}$.

\bmhead{Projector.}
To better retain the semantic space of VLP model, we refer to the design of SideAdapter~\cite{xu2023side} and predict attention biases~$\mathbfcal{B}\in\mathbb{R}^{N \times T \times S \times H \times W}$ to guide VLP model project instance queries $\mathbfcal{Q}$ to instance embeddings $\mathbfcal{E}\in\mathbb{R}^{N \times T \times d} $, where $S$ denotes the number of attention heads. 
Moreover, the category text set is also projected by VLP model to category embeddings~$\mathbfcal{E}^{c}\in\mathbb{R}^{C \times d}$.
We mainly choose CLIP~\cite{radford2021learning} as the VLP model, which is frozen all the time to avoid breaking its alignment space.

\bmhead{Training.}
The whole training process is divided into two stages: (1) The first training stage follows the original implementation of MinVIS~\cite{huang2022minvis} to train the segmentor; (2) In the second stage, we freeze the segmentor and extra introduce the objectives in BTA to optimize the resampler.

\bmhead{Inference.}
During the inference stage, we adopt the window inference strategy~\cite{huang2022minvis} to avoid out-of-memory when processing long videos. Specifically, we set a local window $W$ to slide on the temporal axis of the video for selecting entered frames. Follow the calculation flow of Fig.~\ref{fig:overall_pipeline}, we acquire the instance embeddings of each frame. The average instance embeddings of the head and tail frames~$(\mathbfcal{E}_{1} + \mathbfcal{E}_{T})/2$ are used to align class texts for classifying instances.

\begin{figure}[t]
\centering
\includegraphics[width=1.0\linewidth]{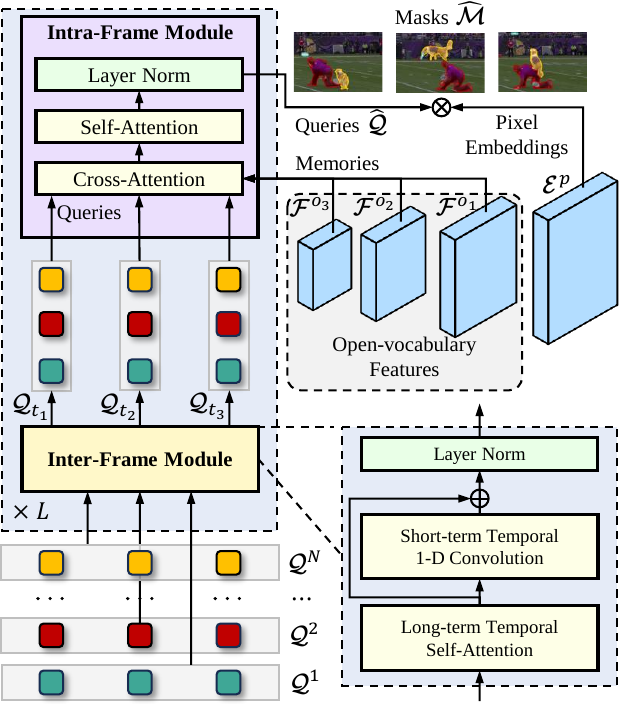} 
\caption{\textbf{Temporal Instance Resampler.} 
The resampler contains inter-frame and intra-frame modules. The former is used to capture long-range and short-range temporal context information. The latter is used to regenerate instance queries via temporal context. With regenerated instance queries~$\mathbfcal{Q}$, we can get new segmentation masks~$\mathbfcal{M}$ with better temporal consistency.
TIR has $L$ times repetition during calculation. In $l$-th time, TIR adopts $\mathbfcal{F}^{o_{[l\%3]}}$ as memories to attend cross-attention of intra-frame module.
}
\vspace{-10pt}
\label{fig:tir}
\end{figure}

\subsection{Temporal Instance Resampler}
\label{sec:tir}

The detailed calculation workflow of TIR is shown in Fig.~\ref{fig:tir}. The TIR is repeated $L$ times for deep supervision. TIR is composed of Inter-frame module and Intra-frame module:

\bmhead{Inter-frame Module.}
Given instance queries $\{\mathbfcal{Q}^1\in\mathbb{R}^{T \times d}, \mathbfcal{Q}^2, \cdots, \mathbfcal{Q}^{N}\}$ extracted by the segmentor, we first input them into self-attention to make instance queries exchange information with each other globally.
Then we input queries into 1-D convolution to aggregate local information of multiple adjacent frames,
where the kernel size and stride of the convolution block are $5$ and $1$, respectively. 
To avoid vanishing the global property, we add a shortcut to replenish long-range context information from self-attention.
The enhanced instance queries are finally normed by layernorm. 

\bmhead{Intra-frame Module.}
The Intra-frame module is essentially a transformer decoder, which can regenerate the instance queries according to the temporal context from Inter-frame module.
After Inter-frame module, we rearrange the shape of instance queries to frame-wise, i.e., $\{\mathbfcal{Q}_{t_1}\in\mathbb{R}^{N \times d}, \mathbfcal{Q}_{t_2}, \mathbfcal{Q}_{t3}\}$. 
To simplify the explanation, we only illustrate the calculation of three frames in Fig.~\ref{fig:tir}. 
The instance queries are input into cross-attention to sample multi-scale pixel features. 
To fully exploit multi-scale features, three scales of open-vocabulary features are circularly utilized by different layers of the TIR. 
Specifically, the $l$-th layer of TIR adopts $\mathbfcal{F}^{o_{[l\%3]}}$ as memories to attend cross-attention with instance queries.
Subsequently, instance queries are processed by self-attention and layernorm.
Finally, regenerated instance queries~$\mathbfcal{Q}$ are used to generate more temporally consistent segmentation masks~$\mathbfcal{M}$ via a dot product with pixel embeddings~$\mathbfcal{E}^p$.

\subsection{Bridge-Text Alignment}
\label{sec:bta}

\begin{figure}[t]
\centering
\begin{subfigure}[b]{0.23\textwidth}
    \centering
    \includegraphics[width=1.0\textwidth]{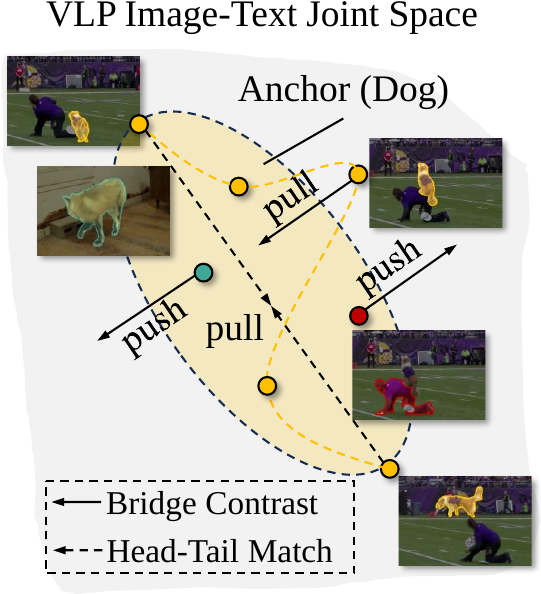}
    \caption{}
    \label{fig:bta_b2b}
\end{subfigure}
\hfill
\begin{subfigure}[b]{0.23\textwidth}
    \centering
    \includegraphics[width=1.0\textwidth]{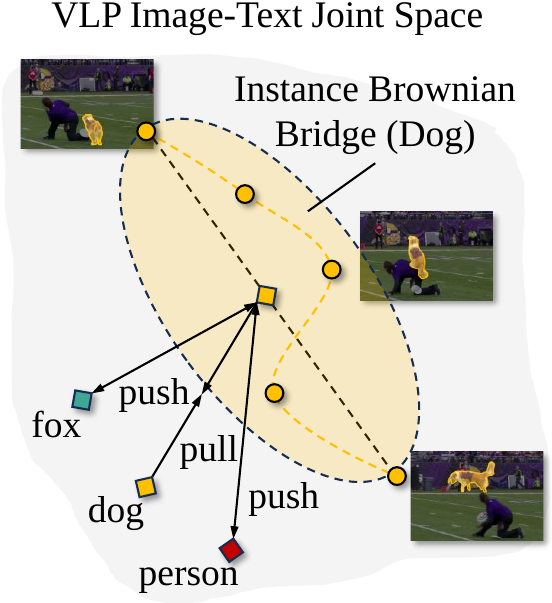}
    \caption{}
    \label{fig:bta_b2t}
\end{subfigure}
\caption{\textbf{Bridge-Text Alignment.} BTA serves as a training mechanism and can be divided into two steps: (1) Linking instances spanning multiple frames as a Brownian bridge; (2) Aligning instance Brownian bridge to class text. The first step is implemented via (a) Head-Tail Matching \& Bridge Contrastive losses. The second step is achieved by (b) Bridge-Text Contrastive loss.
}
\label{fig:bta}
\vspace{-10pt}
\end{figure}

BTA serves as a training mechanism and can be divided into two steps: (1) Linking instance as Brownian bridge; (2) Aligning instance Brownian bridge to class text.
In the former step, we design \textbf{Head-Tail Matching}~(Fig.~\ref{fig:bta_b2b}) for constraining bridge width between head and tail and design \textbf{Bridge Contrastive}~(Fig.~\ref{fig:bta_b2b}) for molding independent frame-level instance features to follow the distribution of Brownian bridge. 
In the latter step, we design \textbf{Bridge-Text Contrastive}~(Fig.~\ref{fig:bta_b2t}) for making instance Brownian bridge align with correct class label text.

Formally, given multiple sampled frames of a video $\mathbfcal{V} = \{I_s, I_{s+1}, ..., I_e\}$, we construct data batch by randomly sampling one frames as triplets~$\{I_{s}, I_{t}, I_{e}\}$, where $1 \leq s < t < e \leq T$. 
After the processing of the segmentor and TIR, we can acquire $B$ batches instance queries~$\mathbfcal{Q}\in\mathbb{R}^{B \times N \times 3\times d}$ and flatten the batch dimension to $\mathbfcal{Q}\in\mathbb{R}^{B \cdot N \times 3\times d}$. 
$\mathbfcal{Q}$ are then projected to instance embeddings~$\mathbfcal{E}\in\mathbb{R}^{B \cdot N \times 3 \times d}$. $B\cdot N$ is set as batch dimension.

\bmhead{Head-Tail Matching.}
Extract head~$\mathbfcal{E}_{s}$ and tail~$\mathbfcal{E}_{e}$ instance embeddings, we adopt a hinge loss to keep the distance between head and tail embeddings in a reasonable range:
\begin{align}
    \mathcal{L}_{htm} &= \mathrm{max}\left(0, \Delta - \mathbfcal{E}_{s} \cdot \mathbfcal{E}_e\right), \\
                      &= \mathrm{ReLU}\left(\Delta - \mathbfcal{E}_{s} \cdot \mathbfcal{E}_e\right),
\end{align}
where the embeddings is processed by $\ell_2$ normalize, $\mathrm{ReLU}$ denotes rectified linear unit activation function, and $\Delta$ denotes the bound value. We convert ReLU to its smooth approximation~\cite{glorot2011deep} to acquiring a gradient-friendly loss:
\begin{equation}
\label{smooth_appro_relu}
\begin{aligned}
    ReLU(z) = \mathrm{max}(0, z) &\approx \mathrm{log}(1 + \mathrm{e}^z).
\end{aligned}
\end{equation}
The final formula of matching loss is:
\begin{equation}
\begin{aligned}
    \mathcal{L}_{htm} = \mathrm{log}\left[1 + \mathrm{e}^{\Delta - \mathbfcal{E}_{s} \cdot \mathbfcal{E}_e}\right]
\label{eq:htm}
\end{aligned}
\end{equation}

\bmhead{Bridge Contrastive.}
The goal of this objective is to make instance embeddings follow the Brownian bridge transition density in Eq.~\ref{eq:brownian_density}, which is ensured via a contrastive objective inspired by~\cite{wang2022language,zhang2023modeling}.
Firstly, we define the contrastive distance measurement between middle states~$\mathbfcal{E}_t$ and the target point in Brownian bridge:
\begin{equation}
d(\mathbfcal{E}^{i}_{s},\mathbfcal{E}^{i}_{t},\mathbfcal{E}^{i}_{e}) = -\frac{1}{2\sigma^{2}}\left\Vert\mathbfcal{E}^{i}_t - (1 - \beta)\mathbfcal{E}^{i}_s - \beta\mathbfcal{E}^{i}_e\right\Vert^{2}_{2},
\end{equation}
where $\sigma^{2}$ is the variance in Eq.~\ref{eq:brownian_density}: $\frac{(t - s)(e - t)}{(e - s)}$, $\beta$ is equal to $\frac{t-s}{e-s}$ and $i$ is the instance index at batch axis, i.e. $i\in [0, B\cdot N-1]$.
Suppose that we set an instance embedding as anchor~$\mathbfcal{E}^{a}$. 
To engrave the Brownian bridge of the anchor instance, the objective push away those negative middle states~$\{\mathbfcal{E}^{i}_{t} | i\neq a\}$ which does not belong to anchor instance and pull the middle state of anchor instance to close the inner point of Brownian bridge. 
As described in previous works about contrastive learning~\cite{robinson2021contrastive}, we should pay more attention to pushing away those informative negative states so we introduce top-K strategy to selecting these confusing middle state as negative set $\mathbfcal{N}$:
\begin{equation}
    \mathbfcal{N}^{a} = \{\mathbfcal{E}^{j}_{t}|j \in \mathrm{topK}(d(\mathbfcal{E}^{a}_{s},\mathbfcal{E}^{j}_{t},\mathbfcal{E}^{a}_{e})), j\neq a\},
\end{equation}
\begin{equation}
    \mathcal{L}^{a}_{bc} = -\mathrm{log}{
    \frac{
    e^{d(\mathbfcal{E}^{a}_{s},\mathbfcal{E}^{a}_{t},\mathbfcal{E}^{a}_{e})}
    }
        {
    e^{d(\mathbfcal{E}^{a}_{s},\mathbfcal{E}^{a}_{t},\mathbfcal{E}^{a}_{e})} + 
    \sum\limits_{\mathbfcal{E}^{j}_{t}\in\mathbfcal{N}^{a}}e^{d(\mathbfcal{E}^{j}_{s},\mathbfcal{E}^{j}_{t},\mathbfcal{E}^{j}_{e})}
    }
    },
\label{eq:bc}
\end{equation}
where $K$ is regularly set to $5$ in our experiments.

\begin{table*}[t]
\centering                         
\renewcommand{\arraystretch}{1.35} 
\setlength{\tabcolsep}{2.5mm}      
\footnotesize                      
\aboverulesep=0ex  
\belowrulesep=0ex
\begin{tabular}{z{80}|x{40}|x{35}|x{35}|x{35}x{35}x{35}|x{35}|x{35}}
\noalign{\hrule height 1.5pt}
\multicolumn{1}{c|}{\multirow{2}{*}{Method}} & \multicolumn{1}{c|}{\multirow{2}{*}{Vocabulary}} & \multicolumn{1}{c|}{\multirow{2}{*}{YTVIS-19}} & \multicolumn{1}{c|}{\multirow{2}{*}{YTVIS-21}} & \multicolumn{3}{c|}{BURST} & \multirow{2}{*}{OVIS} & \multirow{2}{*}{LV-VIS} \\
\cmidrule{5-7}
& & &  & all & common & uncommon &  & \\
\hline
\rowcolor{gray!20}
\multicolumn{9}{c}{\it{Close-vocabulary}}\\
\hline
Mask2Former~\pub{Arxiv2021}~\cite{cheng2021mask2former} &  - & 46.40 & 40.60 & - & - & - & 17.30 & -\\
MinVIS~\pub{Neurips2022}~\cite{huang2022minvis}         &  - & 47.40 & 44.20 & - & - & - & 25.00 & -\\
IDOL~\pub{ECCV2022}~\cite{wu2022defense}                &  - & 49.50 & 43.90 & - & - & - & 28.20 & -\\
SeqFormer~\pub{ECCV2022}~\cite{wu2022seqformer}         &  - & 47.40 & 40.50 & - & - & - & 15.10 & -\\
ViTA~\pub{Neurips2022}~\cite{heo2022vita}               &  - & 49.80 & 45.70 & - & - & - & 19.60 & -\\
GenVIS~\pub{CVPR2023}~\cite{heo2023generalized}         &  - & 50.00 & 47.10 & - & - & - & 37.80 & -\\
\hline
\rowcolor{gray!20}
\multicolumn{9}{c}{\it{Open-vocabulary}}\\
\hline
Detic-S~\pub{ECCV2022}{\cite{zhou2022detic}}    &  L-1203     & 14.60  & 12.70 & 1.90 & -    & -    & 6.70 & 6.50 \\
Detic-O~\pub{ECCV2022}{\cite{zhou2022detic}}    &  L-1203     & 17.90  & 16.70 & 2.70 & -    & -    & 9.00 & 7.70 \\
OV2Seg~\pub{ICCV2023}{\cite{wang2023towards}}   &  L-1203     & 27.20  & 23.60 & 3.70 & -    & -    & 11.2 & 14.20\\
OpenVIS~\pub{Arxiv2023}{\cite{guo2023openvis}}  &  Y-40       & 36.18  & -     & 3.48 & 5.81 & 3.00 & -    & -\\
\hline
\rowcolor{aliceblue!60} $^{*}$OpenVIS &  Y-40  & 23.22 & 17.35 & 3.42 & 6.30 & 2.71 & OOM                     & 9.23  \\
\rowcolor{aliceblue!60} $^{*}$OV2Seg  &  Y-40  & 46.23 & 37.68 & 2.13 & 5.87 & 1.20 & \textbf{14.34}          & 2.08  \\
\rowcolor{aliceblue!60} BriVIS &  Y-40  & \textbf{49.66} & \textbf{42.16} & 3.89 & 8.23 & 2.81 & 13.26 & 10.10 \\
\hdashline
\rowcolor{aliceblue!60} $^{*}$OpenVIS &  LV-1196    & 21.27 & 15.19 & 5.30 & 5.71 & 5.20 & OOM   & 14.43 \\
\rowcolor{aliceblue!60} $^{*}$OV2Seg  &  LV-1196    & 31.48 & 27.56 & 4.97 & 8.38 & 4.12 & 10.03 & 22.27 \\
\rowcolor{aliceblue!60} BriVIS &  LV-1196    & 45.32 & 39.53 & \textbf{7.43} & \textbf{9.53} & \textbf{6.91} & 14.27 & \textbf{27.68} \\
\noalign{\hrule height 1.5pt}
\end{tabular}
\caption{\textbf{Main results} on mainstream VIS datasets. \textbf{Bold} denotes the best performance. ``OOM" denotes out-of-memory on our testing environment. $^{*}$ denotes the results are re-implemented by us. ``Vocabulary" denotes the training dataset of open-vocabulary methods. ``L-1203" denotes the LVIS~\cite{gupta2019lvis} dataset with 1203 categories. ``Y-40" denotes the YouTube-VIS~\cite{yang2019video} dataset with 40 categories. ``LV-1196" denotes the LV-VIS~\cite{wang2023towards} dataset with 1196 categories. ``common" is a subset of the BURST~\cite{athar2023burst} dataset consisting of 78 common object categories, while ``uncommon" denotes another BURST subset composed of 404 uncommon object categories. ``Detic-S"/``Detic-O" are the Detic model~\cite{zhou2022detic} with SORT~\cite{bewley2016simple} or OWTB~\cite{liu2022opening} trackers. The backbone of all models is set to ResNet-50 by default.}
\label{tab:main_res}
\vspace{-10pt}
\end{table*}

\bmhead{Bridge-Text Contrastive.}
To align the instance brownian bridge to class text embeddings~$\mathbfcal{E}^{c}$, according to the goal-conditioned nature, we represent instance via its Brownian bridge center, i.e., the average of the head and tail instance embeddings~$\mathbfcal{E}_{s,t,e}=(\mathbfcal{E}_{s} + \mathbfcal{E}_{e})/2$. Then we adopt a contrastive objective to accomplish this alignment purpose:
\begin{equation}
    \mathcal{L}_{btc} = -\mathrm{log}{
    \frac{
        e^{\mathbfcal{E}_{s,t,e}\cdot\mathbfcal{E}^{c}_{pos}}
    }
        {
        \sum\limits_{k\in\mathbfcal{C}}e^{\mathbfcal{E}_{s,t,e}\cdot\mathbfcal{E}^{c}_{k}}
    }
    },
\label{eq:btc}
\end{equation}
where $\mathbfcal{E}^{c}_{pos}$ denotes the correct category of instance. 

\bmhead{Overall Objective.} 
Associating all of the losses above, we acquire the overall objective of BTA:
\begin{equation}
    \mathcal{L}_{bta} = \frac{1}{B\cdot N}\sum\limits^{B\cdot N}_{a=0}\mathbfcal{L}^{a}_{htm} + \mathbfcal{L}^{a}_{bc} + \mathbfcal{L}^{a}_{btc}.
    \label{eq:bta}
\end{equation}

\section{Experiments}

\subsection{Setup}

\bmhead{Evaluation Protocol.}
We refer to the challenge ``Cross-Dataset" settings of open-vocabulary semantic segmentation~\cite{xu2022simple} to regularize the evaluation protocol of OVVIS.
Previous OVVIS training dataset is limited to small-vocabulary~(Youtube-VIS only contains 40 categories) and granularity inconsistency~(LVIS only contains image-level training materials), which is not suitable for fully exhibiting the open-vocabulary ability of OVVIS.
Therefore, we adopt LV-VIS~(1196 categories) as training dataset and use BURST~(482 categories) as evaluating dataset to sufficiently learn or judge vocabulary transfer ability.
Since the original results of OpenVIS~\cite{guo2023openvis} are trained on Youtube-VIS 2019 so we propose an extra evaluation protocol to use Youtube-VIS 2019 as the training dataset.
Under the two settings described above, we reimplement previous OVVIS methods for comparison.
As for the metric, we use mean Average Precision~(mAP) for evaluation~\cite{yang2019video}.

\bmhead{Implementation Details.}
Following the Mask2former-VIS~\cite{cheng2021mask2former}, We resize the shorter spatial side of video frames to either 360 or 480 and adopt a random horizontal flip strategy.
In the first training stage, the number of sampled frames $T$ is set to $1$. We train our models for 12k iterations
with a batch size of 8 in this stage. Furthermore, we initial the model with the weights of Mask2Former~\cite{cheng2022masked} pretrained
on COCO~\cite{lin2014microsoft} instance segmentation dataset.
In the second stage, we sample $T=5$ frames from videos. We still train models for 12k iterations with a batch size of 8.
AdamW \cite{loshchilov2018decoupled} is adopted as our optimizer, and the learning rate and weight decay are set to 1e-4 and 5e-2.
The learning rate is scaled by a decay factor of $0.1$ at the $10$k iterations of two training stages.
The number of queries $N$ is set to 100.
The bound value $\Delta$ is set to 0.5 by default.
We adopt the ViT-B/16 version of CLIP.
When acquiring the class embedding, we input text in CLIP with prompt templates, e.g., "a photo of \{category\}". To boost open-vocabulary performance, we ensemble 14 text prompt templates from ViLD~\cite{gu2022openvocabulary}.  

\begin{figure}[t]
\centering
\begin{subfigure}[b]{0.23\textwidth}
    \centering
    \includegraphics[width=1.0\textwidth]{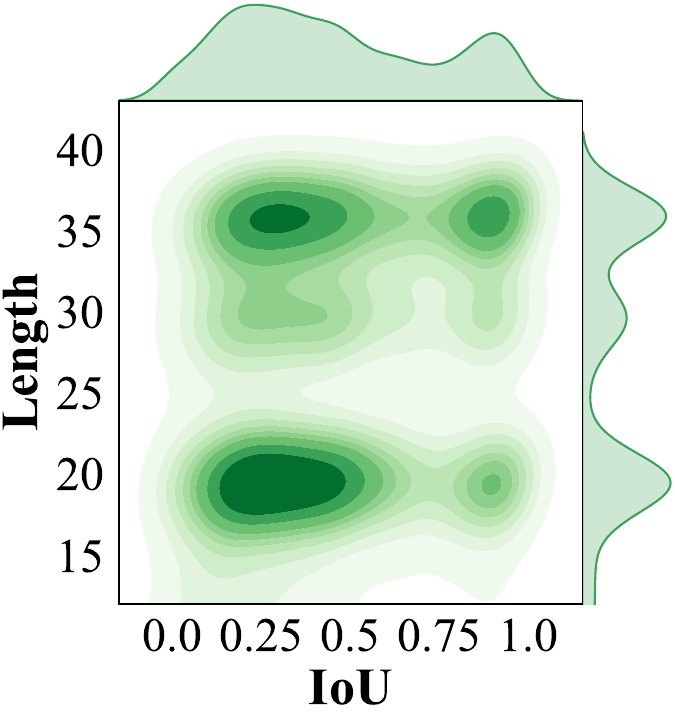}
    \caption{Baseline}
    \label{fig:len_iou_dist_dire}
\end{subfigure}
\hfill
\begin{subfigure}[b]{0.23\textwidth}
    \centering
    \includegraphics[width=1.0\textwidth]{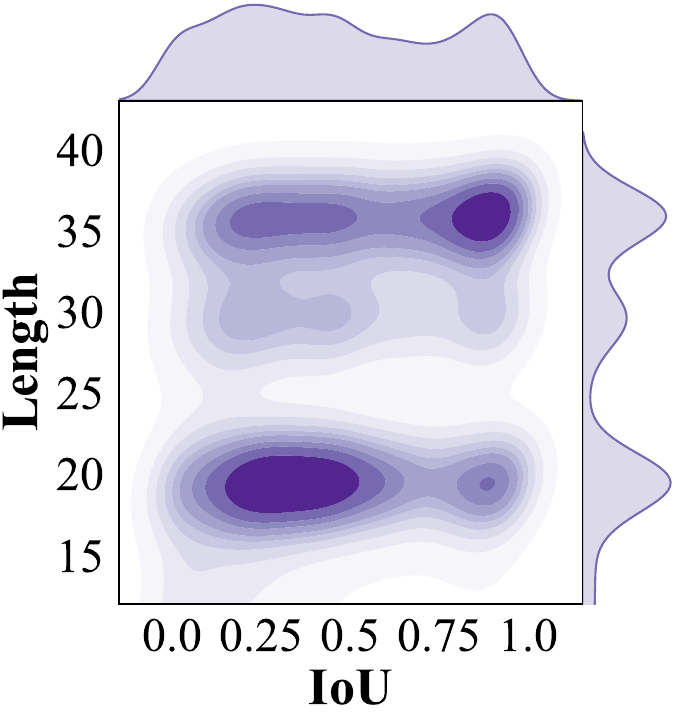}
    \caption{BriVIS}
    \label{fig:len_iou_dist_brow}
\end{subfigure}
\vspace{-5pt}
\caption{\textbf{Correlation between IoU and Video Instance Length} of (a) Baseline and (b) BriVIS~(ours), which is based on Youtube-VIS 2019 train split. The darker area indicates more samples are of the corresponding IoU value and video instance length.
}
\vspace{-10pt}
\label{fig:len_iou_dist}
\end{figure}

\begin{table}
\centering                         
\renewcommand{\arraystretch}{1.4}  
\setlength{\tabcolsep}{1.6mm}      
\footnotesize                      
\begin{tabular}{x{35}x{35}|y{40}y{40}y{40}}
\noalign{\hrule height 1.5pt}
TIR (Fig.~\ref{fig:tir}) & $\mathbfcal{L}_{bta}$ (Fig.~\ref{fig:bta}) & \multicolumn{1}{c}{{mAP$_{a}$}} & \multicolumn{1}{c}{{mAP$_{c}$}} & \multicolumn{1}{c}{{mAP$_{u}$}} \\
\hline
       &        & 5.82 & 9.44 & 4.92 \\
\cmark &        & 6.53~\increase{0.71}  & 9.49~\increase{0.05} & 5.78~\increase{0.86} \\
\rowcolor{aliceblue}
\cmark & \cmark & \textbf{7.43~\increase{1.61}} & \textbf{9.53~\increase{0.09}} & \textbf{6.91~\increase{1.99}}\\
\noalign{\hrule height 1.5pt}
\end{tabular}
\caption{\textbf{Ablation Study} of our designs, i.e., TIR~(Fig.~\ref{fig:tir}) and BTA~($\mathbfcal{L}_{bta}$, Fig.~\ref{fig:bta}). mAP$_{a}$, mAP$_{c}$, and mAP$_{u}$ denote the average AP metric of all, common, and uncommon BURST categories.}
\label{tab:main_abl}
\vspace{-10pt}
\end{table}

\subsection{Main Results}

In Tab.~\ref{tab:main_res}, we compare our BriVIS to previous representative VIS and OVVIS methods, respectively. 

\bmhead{Comparison to Open-vocabulary methods.}
In the ``Open-vocabulary" part of Tab.~\ref{tab:main_res}, we compare our method to OVVIS methods. 
Under evaluation protocol based on LV-VIS training set~(``LV-1196"), our BriVIS improves previous OVVIS methods by a clear margin.
Specifically, compared to previous OVVIS SOTA, our method exhibits 149.49\% performance improvement on the challenge large-vocabulary BURST VIS dataset~(7.43 vs.~4.97 mAP) and increases 13.84, 11.97, and 4.24 mAP performance on Youtube-VIS 2019, Youtube-VIS 2021, and OVIS, demonstrating the superior open-vocabulary ability of our method. 

\bmhead{Comparison to Close-vocabulary methods.}
When comparing BriVIS under ``Y-40" settings to regular VIS methods in the ``Close-vocabulary" part of Tab.~\ref{tab:main_res}, ``inductive" results~(training and evaluating on the same dataset) of our method are nearly equivalent to GenVIS on Youtube-VIS 2019~(49.66 vs.~50.00 mAP) and ``transductive" results~(training and evaluating on the different dataset) of our method are also competitive compared to GenVIS on Youtube-VIS 2021 and OVIS.
These results mean our BriVIS can also serve as an effective VIS method. 

\subsection{Quantitative Analysis}
\label{sec:quant_analysis}

BURST categories can be divided into three folds: ``all", ``common", and ``uncommon", we denote the performance of these three parts as mAP$_{a}$, mAP$_{c}$, and mAP$_{u}$.

\bmhead{Main Ablation.}
To check the effectiveness of our proposed TIR and BTA, we ablate these two modules from OpenBridgeVIS in Tab.~\ref{tab:main_abl}. 
The introduction of TIR enables temporal interaction between independent frame-level instance queries to improve the baseline by +0.70 mAP$_{a}$, +0.05 mAP$_{c}$, and +0.86 mAP$_{u}$. 
The BTA leverages the temporal interaction to guide instance queries spanning video to follow Brownian bridge, which further improves baseline with TIR by +1.61 mAP$_{a}$, +0.09 mAP$_{c}$, and +1.99 mAP$_{u}$.
In summary, the ablation results in Tab.~\ref{tab:main_abl} justify the effectiveness of the proposed TIR and BTA.

\begin{figure}[t]
\centering
\includegraphics[width=1.0\linewidth]{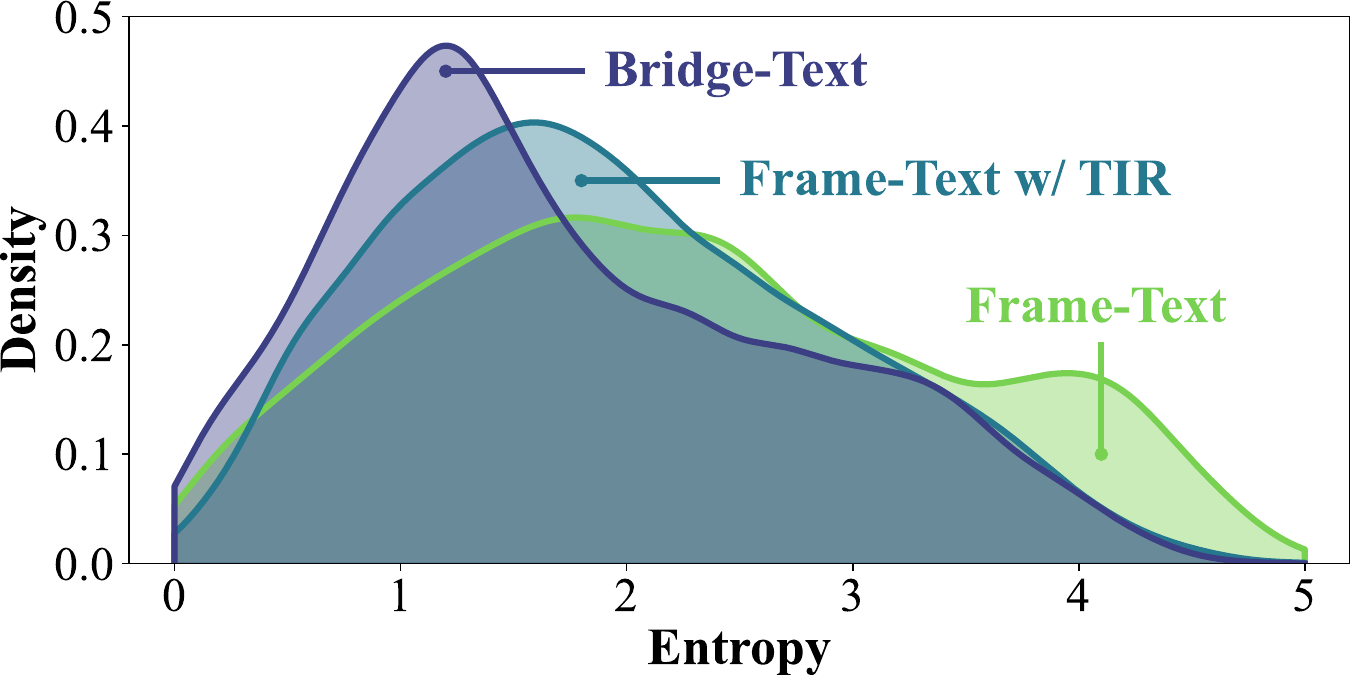} 
\caption{\textbf{Classification Entropy Distribution} of Baseline~(Frame-Text), Baseline w/ TIR~(Frame-Text w/ TIR), BriVIS~(Bridge-Text), which are corresponding to the module combinations in Tab.~\ref{tab:main_abl}. The entropy is calculated based on alignment scores between videos and class texts.
}
\vspace{-10pt}
\label{fig:entro_dist}
\end{figure}

\begin{table}
\centering                         
\renewcommand{\arraystretch}{1.4}  
\setlength{\tabcolsep}{1.6mm}      
\footnotesize                      
\begin{tabular}{x{35}x{35}|y{40}y{40}y{40}}
\noalign{\hrule height 1.5pt}
TIR (Fig.~\ref{fig:tir}) & $\mathbfcal{L}_{bta}$ (Fig.~\ref{fig:bta}) & \multicolumn{1}{c}{{DetA}} & \multicolumn{1}{c}{{AssA}} & \multicolumn{1}{c}{{HOTA}} \\
\hline
       &        & 10.32 & 18.65 & 13.01 \\
\cmark &        & 10.42~\increase{0.10}  & 19.32~\increase{0.67} & 13.42~\increase{0.41} \\
\rowcolor{aliceblue}
\cmark & \cmark & \textbf{10.55~\increase{0.23}} & \textbf{20.47~\increase{1.82}} & \textbf{13.89~\increase{0.88}} \\
\noalign{\hrule height 1.5pt}
\end{tabular}
\caption{\textbf{Tracking Performance} of different combinations in Tab.~\ref{tab:main_abl}. DetA and AssA focus on measuring detection and association, and HOTA is a trade-off metric between DetA and AssA~\cite{luiten2021hota}.}
\label{tab:track_check}
\vspace{-10pt}
\end{table}

\begin{figure*}[t]
\centering
\includegraphics[width=1.0\linewidth]{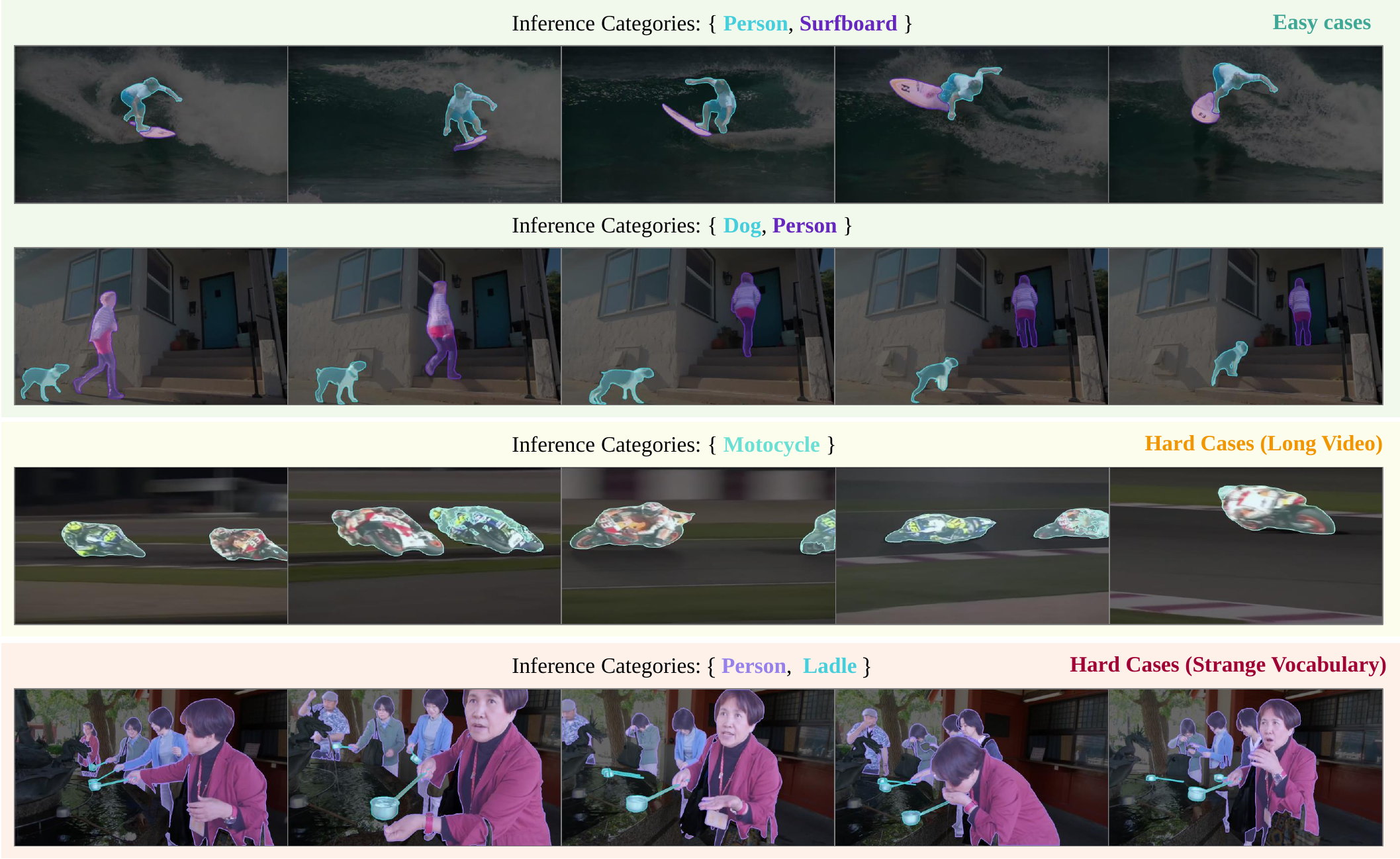} 
\caption{\textbf{Qualitative results of different cases.} We select easy cases in normal scenarios, hard cases in long video scenarios, and hard cases in strange vocabulary scenarios to verify the effectiveness of ours.
}
\label{fig:main_qual}
\vspace{-8pt}
\end{figure*}

\bmhead{How does video length affect the effectiveness?}
To check the scalability of our method for processing instances spanning different frame numbers, we count the ``Length-IoU" density map of the baseline method and our openBridgeVIS in Fig.~\ref{fig:len_iou_dist_dire} and Fig.~\ref{fig:len_iou_dist_brow}, where ``Length" denotes how many frames the instance spans and ``IoU" is the Intersection-over-Union of temporal segmentation masks between matched prediction instance and ground truth instance.
The baseline method and our OpenBridgeVIS correspond to the second row and the last row of Tab.~\ref{tab:main_abl}.
Comparing the two figures, we can find that the top right corner of Fig.~\ref{fig:len_iou_dist_brow} is darker than Fig.~\ref{fig:len_iou_dist_dire}, which means our OpenBridgeVIS prompts better robustness to hard instances with large time span.
Therefore, the introduction of Brownian bridge modeling in our method is meaningful for processing long videos or instances spanning a large number of frames.

\bmhead{The significance of bridge for tracking.}
We re-evaluate those model combinations of Tab.~\ref{tab:main_abl} via multi-object tracking metric, i.e., DetA, AssA, and HOTA~\cite{luiten2021hota}. DetA and AssA respectively concentrate on measuring the detection effectiveness and the association consistency. HOTA is the trade-off metric of DetA and AssA.
As shown in Tab.~\ref{tab:track_check}, our TIR and BTA mainly improve AssA performance of baseline~(e.g., TIR improves baseline +0.67\% AssA, TIR w/ BTA provides a further +1.15\% AssA improvement), verifying that our method design can boost the tracking process of the online video instance segmentor.

\bmhead{The significance of bridge for precise semantic descriptor.}
We analyze the semantic precision of different alignment methods via classification entropy of alignment score. 
Larger entropy denotes the semantic description is more ambiguous, and vice versa. 
As shown in Fig.~\ref{fig:entro_dist}, we plot the classification entropy distribution of Frame-Text~(Baseline), Frame-Text w/ TIR~(Baseline w/ TIR), and Bridge-Text~(BriVIS).
The entropy of the Frame-Text alignment is reduced after the introduction of TIR and is further decreased by converting Frame-Text to Bridge-Text alignment, which means that building connections between independent frame-level instance features and adopting Brownian bridge to guide the connections is significant for constructing more robust and precise alignment between instances and class texts under temporal environment. 

\subsection{Qualitative Analysis}

As described in Quantitative Analysis~(Sec.~\ref{sec:quant_analysis}), the method with bridge-text alignment~(BriVIS) is more capable of processing long video instances~(Fig.~\ref{fig:len_iou_dist}) and provides more robust alignment between video instances and class texts~(Fig.~\ref{fig:entro_dist}) than baseline.
To qualitatively verify this point, we select some easy and hard cases to illustrate the temporal segmentation performance. 
Fig.~\textcolor{red}{\ref{fig:main_qual}} shows that our method generates high-quality spatial-temporal segmentation masks under both easy cases and hard cases, which further justifies the significance of our bridge modeling.

\section{Conclusion}

In this paper, we propose BriVIS to support considering instance dynamics when using the image-text pretraining model to perform open-vocabulary recognition.
Specifically, the BriVIS links independent frame-level instance features as Brownian bridge and aligns the bridge with text in image-text pretraining space for recognizing instances.
The key of our method is to sublimate the granularity of instance descriptor from frame to bridge. 
Bridge-level instance representation is able to carry more context information, especially for instance movement context.
Therefore, our method exhibits a more robust and precise alignment between instance and class texts, verified by classification entropy distribution statistics.
In addition, according to the multi-object tracking performance and the effectiveness of processing long video, the bridge-level instance feature further demonstrates its significance.

{
    \small
    \bibliographystyle{ieeenat_fullname}
    \bibliography{main}

\begin{thebibliography}{67}
\providecommand{\natexlab}[1]{#1}
\providecommand{\url}[1]{\texttt{#1}}
\expandafter\ifx\csname urlstyle\endcsname\relax
  \providecommand{\doi}[1]{doi: #1}\else
  \providecommand{\doi}{doi: \begingroup \urlstyle{rm}\Url}\fi

\bibitem[Athar et~al.(2023)Athar, Luiten, Voigtlaender, Khurana, Dave, Leibe, and Ramanan]{athar2023burst}
Ali Athar, Jonathon Luiten, Paul Voigtlaender, Tarasha Khurana, Achal Dave, Bastian Leibe, and Deva Ramanan.
\newblock Burst: A benchmark for unifying object recognition, segmentation and tracking in video.
\newblock In \emph{Proceedings of the IEEE/CVF Winter Conference on Applications of Computer Vision}, pages 1674--1683, 2023.

\bibitem[Bewley et~al.(2016)Bewley, Ge, Ott, Ramos, and Upcroft]{bewley2016simple}
Alex Bewley, Zongyuan Ge, Lionel Ott, Fabio Ramos, and Ben Upcroft.
\newblock Simple online and realtime tracking.
\newblock In \emph{Proceedings of the IEEE international conference on image processing}, pages 3464--3468. IEEE, 2016.

\bibitem[Bucher et~al.(2019)Bucher, Vu, Cord, and P{\'e}rez]{bucher2019zero}
Maxime Bucher, Tuan-Hung Vu, Matthieu Cord, and Patrick P{\'e}rez.
\newblock Zero-shot semantic segmentation.
\newblock \emph{Advances in Neural Information Processing Systems}, 32, 2019.

\bibitem[Carion et~al.(2020)Carion, Massa, Synnaeve, Usunier, Kirillov, and Zagoruyko]{carion2020end}
Nicolas Carion, Francisco Massa, Gabriel Synnaeve, Nicolas Usunier, Alexander Kirillov, and Sergey Zagoruyko.
\newblock End-to-end object detection with transformers.
\newblock In \emph{Proceedings of the European Conference on Computer Vision}, pages 213--229. Springer, 2020.

\bibitem[Cheng et~al.(2021)Cheng, Choudhuri, Misra, Kirillov, Girdhar, and Schwing]{cheng2021mask2former}
Bowen Cheng, Anwesa Choudhuri, Ishan Misra, Alexander Kirillov, Rohit Girdhar, and Alexander~G Schwing.
\newblock Mask2former for video instance segmentation.
\newblock \emph{arXiv preprint arXiv:2112.10764}, 2021.

\bibitem[Cheng et~al.(2022)Cheng, Misra, Schwing, Kirillov, and Girdhar]{cheng2022masked}
Bowen Cheng, Ishan Misra, Alexander~G Schwing, Alexander Kirillov, and Rohit Girdhar.
\newblock Masked-attention mask transformer for universal image segmentation.
\newblock In \emph{Proceedings of the IEEE/CVF conference on computer vision and pattern recognition}, pages 1290--1299, 2022.

\bibitem[Dave et~al.(2020)Dave, Khurana, Tokmakov, Schmid, and Ramanan]{dave2020tao}
Achal Dave, Tarasha Khurana, Pavel Tokmakov, Cordelia Schmid, and Deva Ramanan.
\newblock Tao: A large-scale benchmark for tracking any object.
\newblock In \emph{Proceedings of the European conference on computer vision}, pages 436--454. Springer, 2020.

\bibitem[Ding et~al.(2022)Ding, Hui, Huang, Wei, Han, and Liu]{ding2022language}
Zihan Ding, Tianrui Hui, Junshi Huang, Xiaoming Wei, Jizhong Han, and Si Liu.
\newblock Language-bridged spatial-temporal interaction for referring video object segmentation.
\newblock In \emph{Proceedings of the IEEE/CVF Conference on Computer Vision and Pattern Recognition}, pages 4964--4973, 2022.

\bibitem[Dosovitskiy et~al.(2020)Dosovitskiy, Beyer, Kolesnikov, Weissenborn, Zhai, Unterthiner, Dehghani, Minderer, Heigold, Gelly, et~al.]{dosovitskiy2020image}
Alexey Dosovitskiy, Lucas Beyer, Alexander Kolesnikov, Dirk Weissenborn, Xiaohua Zhai, Thomas Unterthiner, Mostafa Dehghani, Matthias Minderer, Georg Heigold, Sylvain Gelly, et~al.
\newblock An image is worth 16x16 words: Transformers for image recognition at scale.
\newblock \emph{arXiv preprint arXiv:2010.11929}, 2020.

\bibitem[Ghiasi et~al.(2022)Ghiasi, Gu, Cui, and Lin]{ghiasi2022scaling}
Golnaz Ghiasi, Xiuye Gu, Yin Cui, and Tsung-Yi Lin.
\newblock Scaling open-vocabulary image segmentation with image-level labels.
\newblock In \emph{European Conference on Computer Vision}, pages 540--557. Springer, 2022.

\bibitem[Glorot et~al.(2011)Glorot, Bordes, and Bengio]{glorot2011deep}
Xavier Glorot, Antoine Bordes, and Yoshua Bengio.
\newblock Deep sparse rectifier neural networks.
\newblock In \emph{Proceedings of the fourteenth international conference on artificial intelligence and statistics}, pages 315--323. JMLR Workshop and Conference Proceedings, 2011.

\bibitem[Gu et~al.(2021)Gu, Lin, Kuo, and Cui]{gu2021open}
Xiuye Gu, Tsung-Yi Lin, Weicheng Kuo, and Yin Cui.
\newblock Open-vocabulary object detection via vision and language knowledge distillation.
\newblock In \emph{Proceedings of the International Conference on Learning Representations}, 2021.

\bibitem[Gu et~al.(2022)Gu, Lin, Kuo, and Cui]{gu2022openvocabulary}
Xiuye Gu, Tsung-Yi Lin, Weicheng Kuo, and Yin Cui.
\newblock Open-vocabulary object detection via vision and language knowledge distillation.
\newblock In \emph{International Conference on Learning Representations}, 2022.

\bibitem[Guo et~al.(2023)Guo, Huang, He, Liu, Xiao, Chen, and Zhang]{guo2023openvis}
Pinxue Guo, Tony Huang, Peiyang He, Xuefeng Liu, Tianjun Xiao, Zhaoyu Chen, and Wenqiang Zhang.
\newblock Openvis: Open-vocabulary video instance segmentation.
\newblock \emph{arXiv preprint arXiv:2305.16835}, 2023.

\bibitem[Gupta et~al.(2019)Gupta, Dollar, and Girshick]{gupta2019lvis}
Agrim Gupta, Piotr Dollar, and Ross Girshick.
\newblock Lvis: A dataset for large vocabulary instance segmentation.
\newblock In \emph{Proceedings of the IEEE/CVF Conference on Computer Vision and Pattern Recognition}, pages 5356--5364, 2019.

\bibitem[Han et~al.(2023)Han, Zhong, Li, Han, and Ma]{han2023open}
Cong Han, Yujie Zhong, Dengjie Li, Kai Han, and Lin Ma.
\newblock Open-vocabulary semantic segmentation with decoupled one-pass network.
\newblock In \emph{Proceedings of the IEEE/CVF International Conference on Computer Vision}, pages 1086--1096, 2023.

\bibitem[Heo et~al.(2022)Heo, Hwang, Oh, Lee, and Kim]{heo2022vita}
Miran Heo, Sukjun Hwang, Seoung~Wug Oh, Joon-Young Lee, and Seon~Joo Kim.
\newblock Vita: Video instance segmentation via object token association.
\newblock 2022.

\bibitem[Heo et~al.(2023)Heo, Hwang, Hyun, Kim, Oh, Lee, and Kim]{heo2023generalized}
Miran Heo, Sukjun Hwang, Jeongseok Hyun, Hanjung Kim, Seoung~Wug Oh, Joon-Young Lee, and Seon~Joo Kim.
\newblock A generalized framework for video instance segmentation.
\newblock In \emph{Proceedings of the IEEE/CVF Conference on Computer Vision and Pattern Recognition}, pages 14623--14632, 2023.

\bibitem[Hou et~al.(2021)Hou, Chang, Ma, Huang, and Shan]{hou2021bicnet}
Ruibing Hou, Hong Chang, Bingpeng Ma, Rui Huang, and Shiguang Shan.
\newblock Bicnet-tks: Learning efficient spatial-temporal representation for video person re-identification.
\newblock In \emph{Proceedings of the IEEE/CVF conference on computer vision and pattern recognition}, pages 2014--2023, 2021.

\bibitem[Huang et~al.(2022)Huang, Yu, and Anandkumar]{huang2022minvis}
De-An Huang, Zhiding Yu, and Anima Anandkumar.
\newblock Minvis: A minimal video instance segmentation framework without video-based training.
\newblock \emph{arXiv preprint arXiv:2208.02245}, 2022.

\bibitem[Hui et~al.(2021)Hui, Huang, Liu, Ding, Li, Wang, Han, and Wang]{hui2021collaborative}
Tianrui Hui, Shaofei Huang, Si Liu, Zihan Ding, Guanbin Li, Wenguan Wang, Jizhong Han, and Fei Wang.
\newblock Collaborative spatial-temporal modeling for language-queried video actor segmentation.
\newblock In \emph{Proceedings of the IEEE/CVF Conference on Computer Vision and Pattern Recognition}, pages 4187--4196, 2021.

\bibitem[Hui et~al.(2023)Hui, Liu, Ding, Huang, Li, Wang, Liu, and Han]{hui2023language}
Tianrui Hui, Si Liu, Zihan Ding, Shaofei Huang, Guanbin Li, Wenguan Wang, Luoqi Liu, and Jizhong Han.
\newblock Language-aware spatial-temporal collaboration for referring video segmentation.
\newblock \emph{IEEE Transactions on Pattern Analysis and Machine Intelligence}, 2023.

\bibitem[Huynh et~al.(2022)Huynh, Kuen, Lin, Gu, and Elhamifar]{huynh2022open}
Dat Huynh, Jason Kuen, Zhe Lin, Jiuxiang Gu, and Ehsan Elhamifar.
\newblock Open-vocabulary instance segmentation via robust cross-modal pseudo-labeling.
\newblock In \emph{Proceedings of the IEEE/CVF Conference on Computer Vision and Pattern Recognition}, pages 7020--7031, 2022.

\bibitem[Hwang et~al.(2021)Hwang, Heo, Oh, and Kim]{hwang2021video}
Sukjun Hwang, Miran Heo, Seoung~Wug Oh, and Seon~Joo Kim.
\newblock Video instance segmentation using inter-frame communication transformers.
\newblock \emph{Advances in Neural Information Processing Systems}, 34:\penalty0 13352--13363, 2021.

\bibitem[Jia et~al.(2021)Jia, Yang, Xia, Chen, Parekh, Pham, Le, Sung, Li, and Duerig]{jia2021scaling}
Chao Jia, Yinfei Yang, Ye Xia, Yi-Ting Chen, Zarana Parekh, Hieu Pham, Quoc Le, Yun-Hsuan Sung, Zhen Li, and Tom Duerig.
\newblock Scaling up visual and vision-language representation learning with noisy text supervision.
\newblock In \emph{International conference on machine learning}, pages 4904--4916. PMLR, 2021.

\bibitem[Ke et~al.(2022)Ke, Ding, Danelljan, Tai, Tang, and Yu]{ke2022video}
Lei Ke, Henghui Ding, Martin Danelljan, Yu-Wing Tai, Chi-Keung Tang, and Fisher Yu.
\newblock Video mask transfiner for high-quality video instance segmentation.
\newblock In \emph{European Conference on Computer Vision}, pages 731--747. Springer, 2022.

\bibitem[Kuhn(1955)]{kuhn1955hungarian}
Harold~W Kuhn.
\newblock The hungarian method for the assignment problem.
\newblock \emph{Naval research logistics quarterly}, 2\penalty0 (1-2):\penalty0 83--97, 1955.

\bibitem[Li et~al.(2022{\natexlab{a}})Li, Weinberger, Belongie, Koltun, and Ranftl]{li2022languagedriven}
Boyi Li, Kilian~Q Weinberger, Serge Belongie, Vladlen Koltun, and Rene Ranftl.
\newblock Language-driven semantic segmentation.
\newblock In \emph{International Conference on Learning Representations}, 2022{\natexlab{a}}.

\bibitem[Li et~al.(2022{\natexlab{b}})Li, Li, Xiong, and Hoi]{li2022blip}
Junnan Li, Dongxu Li, Caiming Xiong, and Steven Hoi.
\newblock Blip: Bootstrapping language-image pre-training for unified vision-language understanding and generation.
\newblock In \emph{International Conference on Machine Learning}, pages 12888--12900. PMLR, 2022{\natexlab{b}}.

\bibitem[Li et~al.(2022{\natexlab{c}})Li, Zhang, Zhang, Yang, Li, Zhong, Wang, Yuan, Zhang, Hwang, et~al.]{li2022grounded}
Liunian~Harold Li, Pengchuan Zhang, Haotian Zhang, Jianwei Yang, Chunyuan Li, Yiwu Zhong, Lijuan Wang, Lu Yuan, Lei Zhang, Jenq-Neng Hwang, et~al.
\newblock Grounded language-image pre-training.
\newblock In \emph{Proceedings of the IEEE/CVF Conference on Computer Vision and Pattern Recognition}, pages 10965--10975, 2022{\natexlab{c}}.

\bibitem[Li et~al.(2023)Li, Li, Xiang, and Zhang]{li2023mdqe}
Minghan Li, Shuai Li, Wangmeng Xiang, and Lei Zhang.
\newblock Mdqe: Mining discriminative query embeddings to segment occluded instances on challenging videos.
\newblock In \emph{Proceedings of the IEEE/CVF Conference on Computer Vision and Pattern Recognition}, pages 10524--10533, 2023.

\bibitem[Liang et~al.(2023)Liang, Wu, Dai, Li, Zhao, Zhang, Zhang, Vajda, and Marculescu]{liang2023open}
Feng Liang, Bichen Wu, Xiaoliang Dai, Kunpeng Li, Yinan Zhao, Hang Zhang, Peizhao Zhang, Peter Vajda, and Diana Marculescu.
\newblock Open-vocabulary semantic segmentation with mask-adapted clip.
\newblock In \emph{Proceedings of the IEEE/CVF Conference on Computer Vision and Pattern Recognition}, pages 7061--7070, 2023.

\bibitem[Lin et~al.(2014)Lin, Maire, Belongie, Hays, Perona, Ramanan, Doll{\'a}r, and Zitnick]{lin2014microsoft}
Tsung-Yi Lin, Michael Maire, Serge Belongie, James Hays, Pietro Perona, Deva Ramanan, Piotr Doll{\'a}r, and C~Lawrence Zitnick.
\newblock Microsoft coco: Common objects in context.
\newblock In \emph{Proceedings of the European Conference on Computer Vision}, pages 740--755. Springer, 2014.

\bibitem[Liu et~al.(2022)Liu, Zulfikar, Luiten, Dave, Ramanan, Leibe, O{\v{s}}ep, and Leal-Taix{\'e}]{liu2022opening}
Yang Liu, Idil~Esen Zulfikar, Jonathon Luiten, Achal Dave, Deva Ramanan, Bastian Leibe, Aljo{\v{s}}a O{\v{s}}ep, and Laura Leal-Taix{\'e}.
\newblock Opening up open world tracking.
\newblock In \emph{Proceedings of the IEEE/CVF Conference on Computer Vision and Pattern Recognition}, pages 19045--19055, 2022.

\bibitem[Loshchilov and Hutter(2019)]{loshchilov2018decoupled}
Ilya Loshchilov and Frank Hutter.
\newblock Decoupled weight decay regularization.
\newblock In \emph{International Conference on Learning Representations}, 2019.

\bibitem[Luiten et~al.(2021)Luiten, Osep, Dendorfer, Torr, Geiger, Leal-Taix{\'e}, and Leibe]{luiten2021hota}
Jonathon Luiten, Aljosa Osep, Patrick Dendorfer, Philip Torr, Andreas Geiger, Laura Leal-Taix{\'e}, and Bastian Leibe.
\newblock Hota: A higher order metric for evaluating multi-object tracking.
\newblock \emph{International journal of computer vision}, 129:\penalty0 548--578, 2021.

\bibitem[Mikolov et~al.(2013)Mikolov, Sutskever, Chen, Corrado, and Dean]{mikolov2013distributed}
Tomas Mikolov, Ilya Sutskever, Kai Chen, Greg~S Corrado, and Jeff Dean.
\newblock Distributed representations of words and phrases and their compositionality.
\newblock \emph{Advances in neural information processing systems}, 26, 2013.

\bibitem[Miller(1995)]{miller1995wordnet}
George~A Miller.
\newblock Wordnet: a lexical database for english.
\newblock \emph{Communications of the ACM}, 38\penalty0 (11):\penalty0 39--41, 1995.

\bibitem[Qi et~al.(2022)Qi, Gao, Hu, Wang, Liu, Bai, Belongie, Yuille, Torr, and Bai]{qi2022occluded}
Jiyang Qi, Yan Gao, Yao Hu, Xinggang Wang, Xiaoyu Liu, Xiang Bai, Serge Belongie, Alan Yuille, Philip Torr, and Song Bai.
\newblock Occluded video instance segmentation: A benchmark.
\newblock \emph{International Journal of Computer Vision}, 2022.

\bibitem[Radford et~al.(2021)Radford, Kim, Hallacy, Ramesh, Goh, Agarwal, Sastry, Askell, Mishkin, Clark, et~al.]{radford2021learning}
Alec Radford, Jong~Wook Kim, Chris Hallacy, Aditya Ramesh, Gabriel Goh, Sandhini Agarwal, Girish Sastry, Amanda Askell, Pamela Mishkin, Jack Clark, et~al.
\newblock Learning transferable visual models from natural language supervision.
\newblock In \emph{International Conference on Machine Learning}, pages 8748--8763. PMLR, 2021.

\bibitem[Revuz and Yor(2013)]{revuz2013continuous}
Daniel Revuz and Marc Yor.
\newblock \emph{Continuous martingales and Brownian motion}.
\newblock Springer Science \& Business Media, Berlin, 2013.

\bibitem[Robinson et~al.(2021)Robinson, Chuang, Sra, and Jegelka]{robinson2021contrastive}
Joshua~David Robinson, Ching-Yao Chuang, Suvrit Sra, and Stefanie Jegelka.
\newblock Contrastive learning with hard negative samples.
\newblock In \emph{International Conference on Learning Representations}, 2021.

\bibitem[Sun et~al.(2023)Sun, Fang, Wu, Wang, and Cao]{sun2023eva}
Quan Sun, Yuxin Fang, Ledell Wu, Xinlong Wang, and Yue Cao.
\newblock Eva-clip: Improved training techniques for clip at scale.
\newblock \emph{arXiv preprint arXiv:2303.15389}, 2023.

\bibitem[Tu et~al.(2023)Tu, Dai, Wu, Cheng, Hu, and Jiang]{tu2023implicit}
Shuyuan Tu, Qi Dai, Zuxuan Wu, Zhi-Qi Cheng, Han Hu, and Yu-Gang Jiang.
\newblock Implicit temporal modeling with learnable alignment for video recognition.
\newblock \emph{arXiv preprint arXiv:2304.10465}, 2023.

\bibitem[Tu et~al.(2017)Tu, Zhang, Liu, and Yan]{tu2017video}
Yunbin Tu, Xishan Zhang, Bingtao Liu, and Chenggang Yan.
\newblock Video description with spatial-temporal attention.
\newblock In \emph{Proceedings of the 25th ACM international conference on Multimedia}, pages 1014--1022, 2017.

\bibitem[Wang et~al.(2023{\natexlab{a}})Wang, Wang, Yan, Jiang, Tang, Hu, Xie, and Gavves]{wang2023towards}
Haochen Wang, Shuai Wang, Cilin Yan, Xiaolong Jiang, XU Tang, Yao Hu, Weidi Xie, and Efstratios Gavves.
\newblock Towards open-vocabulary video instance segmentation.
\newblock \emph{arXiv preprint arXiv:2304.01715}, 2023{\natexlab{a}}.

\bibitem[Wang et~al.(2023{\natexlab{b}})Wang, Lin, and Li]{wang-etal-2023-dialogue}
Jian Wang, Dongding Lin, and Wenjie Li.
\newblock Dialogue planning via brownian bridge stochastic process for goal-directed proactive dialogue.
\newblock In \emph{Findings of the Association for Computational Linguistics: ACL 2023}, pages 370--387, Toronto, Canada, 2023{\natexlab{b}}. Association for Computational Linguistics.

\bibitem[Wang et~al.(2022)Wang, Durmus, Goodman, and Hashimoto]{wang2022language}
Rose~E Wang, Esin Durmus, Noah Goodman, and Tatsunori Hashimoto.
\newblock Language modeling via stochastic processes.
\newblock In \emph{International Conference on Learning Representations}, 2022.

\bibitem[Wang et~al.(2021)Wang, Xu, Wang, Shen, Cheng, Shen, and Xia]{wang2021end}
Yuqing Wang, Zhaoliang Xu, Xinlong Wang, Chunhua Shen, Baoshan Cheng, Hao Shen, and Huaxia Xia.
\newblock End-to-end video instance segmentation with transformers.
\newblock In \emph{Proceedings of the IEEE/CVF Conference on Computer Vision and Pattern Recognition}, pages 8741--8750, 2021.

\bibitem[Wu et~al.(2022{\natexlab{a}})Wu, Jiang, Bai, Zhang, and Bai]{wu2022seqformer}
Junfeng Wu, Yi Jiang, Song Bai, Wenqing Zhang, and Xiang Bai.
\newblock Seqformer: Sequential transformer for video instance segmentation.
\newblock In \emph{European Conference on Computer Vision}, pages 553--569. Springer, 2022{\natexlab{a}}.

\bibitem[Wu et~al.(2022{\natexlab{b}})Wu, Jiang, Sun, Yuan, and Luo]{wu2022language}
Jiannan Wu, Yi Jiang, Peize Sun, Zehuan Yuan, and Ping Luo.
\newblock Language as queries for referring video object segmentation.
\newblock In \emph{Proceedings of the IEEE/CVF Conference on Computer Vision and Pattern Recognition}, pages 4974--4984, 2022{\natexlab{b}}.

\bibitem[Wu et~al.(2022{\natexlab{c}})Wu, Liu, Jiang, Bai, Yuille, and Bai]{wu2022defense}
Junfeng Wu, Qihao Liu, Yi Jiang, Song Bai, Alan Yuille, and Xiang Bai.
\newblock In defense of online models for video instance segmentation.
\newblock In \emph{European Conference on Computer Vision}, pages 588--605. Springer, 2022{\natexlab{c}}.

\bibitem[Wu et~al.(2022{\natexlab{d}})Wu, Yarram, Liang, Lan, Yuan, Eledath, and Medioni]{wu2022efficient}
Jialian Wu, Sudhir Yarram, Hui Liang, Tian Lan, Junsong Yuan, Jayan Eledath, and Gerard Medioni.
\newblock Efficient video instance segmentation via tracklet query and proposal.
\newblock In \emph{Proceedings of the IEEE/CVF Conference on Computer Vision and Pattern Recognition}, pages 959--968, 2022{\natexlab{d}}.

\bibitem[Xian et~al.(2019)Xian, Choudhury, He, Schiele, and Akata]{xian2019semantic}
Yongqin Xian, Subhabrata Choudhury, Yang He, Bernt Schiele, and Zeynep Akata.
\newblock Semantic projection network for zero-and few-label semantic segmentation.
\newblock In \emph{Proceedings of the IEEE/CVF Conference on Computer Vision and Pattern Recognition}, pages 8256--8265, 2019.

\bibitem[Xu et~al.(2022)Xu, Zhang, Wei, Lin, Cao, Hu, and Bai]{xu2022simple}
Mengde Xu, Zheng Zhang, Fangyun Wei, Yutong Lin, Yue Cao, Han Hu, and Xiang Bai.
\newblock A simple baseline for open-vocabulary semantic segmentation with pre-trained vision-language model.
\newblock In \emph{European Conference on Computer Vision}, pages 736--753. Springer, 2022.

\bibitem[Xu et~al.(2023)Xu, Zhang, Wei, Hu, and Bai]{xu2023side}
Mengde Xu, Zheng Zhang, Fangyun Wei, Han Hu, and Xiang Bai.
\newblock Side adapter network for open-vocabulary semantic segmentation.
\newblock In \emph{Proceedings of the IEEE/CVF Conference on Computer Vision and Pattern Recognition}, pages 2945--2954, 2023.

\bibitem[Yang et~al.(2019)Yang, Fan, and Xu]{yang2019video}
Linjie Yang, Yuchen Fan, and Ning Xu.
\newblock Video instance segmentation.
\newblock In \emph{Proceedings of the IEEE/CVF International Conference on Computer Vision}, pages 5188--5197, 2019.

\bibitem[Yang et~al.(2022)Yang, Wang, Li, Fang, Fang, Liu, Zhao, and Shan]{yang2022temporally}
Shusheng Yang, Xinggang Wang, Yu Li, Yuxin Fang, Jiemin Fang, Wenyu Liu, Xun Zhao, and Ying Shan.
\newblock Temporally efficient vision transformer for video instance segmentation.
\newblock In \emph{Proceedings of the IEEE/CVF Conference on Computer Vision and Pattern Recognition}, pages 2885--2895, 2022.

\bibitem[Yu et~al.(2022)Yu, Wang, Vasudevan, Yeung, Seyedhosseini, and Wu]{yu2022coca}
Jiahui Yu, Zirui Wang, Vijay Vasudevan, Legg Yeung, Mojtaba Seyedhosseini, and Yonghui Wu.
\newblock Coca: Contrastive captioners are image-text foundation models.
\newblock \emph{arXiv preprint arXiv:2205.01917}, 2022.

\bibitem[Yuan et~al.(2021)Yuan, Chen, Chen, Codella, Dai, Gao, Hu, Huang, Li, Li, et~al.]{yuan2021florence}
Lu Yuan, Dongdong Chen, Yi-Ling Chen, Noel Codella, Xiyang Dai, Jianfeng Gao, Houdong Hu, Xuedong Huang, Boxin Li, Chunyuan Li, et~al.
\newblock Florence: A new foundation model for computer vision.
\newblock \emph{arXiv preprint arXiv:2111.11432}, 2021.

\bibitem[Zhan et~al.(2022)Zhan, McKee, and Lazebnik]{zhan2022robust}
Zitong Zhan, Daniel McKee, and Svetlana Lazebnik.
\newblock Robust online video instance segmentation with track queries.
\newblock \emph{arXiv preprint arXiv:2211.09108}, 2022.

\bibitem[Zhang et~al.(2023{\natexlab{a}})Zhang, Liu, Zheng, and Su]{zhang2023modeling}
Heng Zhang, Daqing Liu, Qi Zheng, and Bing Su.
\newblock Modeling video as stochastic processes for fine-grained video representation learning.
\newblock In \emph{Proceedings of the IEEE/CVF Conference on Computer Vision and Pattern Recognition}, pages 2225--2234, 2023{\natexlab{a}}.

\bibitem[Zhang et~al.(2023{\natexlab{b}})Zhang, Tian, Wu, Ji, Wang, Zhang, and Wan]{zhang2023dvis}
Tao Zhang, Xingye Tian, Yu Wu, Shunping Ji, Xuebo Wang, Yuan Zhang, and Pengfei Wan.
\newblock Dvis: Decoupled video instance segmentation framework.
\newblock \emph{arXiv preprint arXiv:2306.03413}, 2023{\natexlab{b}}.

\bibitem[Zhao et~al.(2017)Zhao, Puig, Zhou, Fidler, and Torralba]{zhao2017open}
Hang Zhao, Xavier Puig, Bolei Zhou, Sanja Fidler, and Antonio Torralba.
\newblock Open vocabulary scene parsing.
\newblock In \emph{Proceedings of the IEEE International Conference on Computer Vision}, pages 2002--2010, 2017.

\bibitem[Zheng~Ding(2023)]{ding2023maskclip}
Zhuowen~Tu Zheng~Ding, Jieke~Wang.
\newblock Open-vocabulary universal image segmentation with maskclip.
\newblock In \emph{International Conference on Machine Learning}, 2023.

\bibitem[Zhou et~al.(2022{\natexlab{a}})Zhou, Loy, and Dai]{zhou2022extract}
Chong Zhou, Chen~Change Loy, and Bo Dai.
\newblock Extract free dense labels from clip.
\newblock In \emph{European Conference on Computer Vision}, pages 696--712. Springer, 2022{\natexlab{a}}.

\bibitem[Zhou et~al.(2022{\natexlab{b}})Zhou, Girdhar, Joulin, Kr{\"a}henb{\"u}hl, and Misra]{zhou2022detic}
Xingyi Zhou, Rohit Girdhar, Armand Joulin, Philipp Kr{\"a}henb{\"u}hl, and Ishan Misra.
\newblock Detecting twenty-thousand classes using image-level supervision.
\newblock In \emph{Proceedings of the European Conference on Computer Vision}, 2022{\natexlab{b}}.

\end{thebibliography}
}

\clearpage
\appendix

In this document, we first discuss the limitations of our method~(\S Sec.~\ref{sup_sec:limi}). Then we describe the correlation between instance movement and stochastic process to elaborate on how we model the instance movement context via Brownian bridge in Sec.~``Introduction"~(\S Sec.~\ref{sup_sec:brow}). Subsequently, we explain the design idea of Brownian bridge distance metric~(\S Sec.~\ref{sup_sec:brow_bri_dist}).
In \S Sec.~\ref{sup_sec:dataset}, we describe the details of the selected dataset and analyze the cross-dataset similarity to check the reasonability of our evaluation protocol. Finally, extra ablation results about sub-module design choices and hyperparameters~(\S Sec.~\ref{sup_sec:abla}) are provided.

\section{Limitations}
\label{sup_sec:limi}

In this section, we discuss the limitations of our BriVIS. The limitations can be divided into two folds: 
Firstly, the Brownian bridge modeling requires to access instance features of whole video so BriVIS is designed to follow the offline Video Instance Segmentation paradigm, causing BriVIS hard to process long videos or video streams because of the high computation cost~\cite{wu2022defense}.
Secondly, the clip visual space is not able to accommodate temporal context. Therefore, the Brownian bridge modeling based on clip frame features can only implicitly model the temporal context, i.e., aligning frame features in parallel with an aim to involve motion and action clues~\cite{tu2023implicit}.
Such a modeling method is difficult to extract profound reasoning temporal context and is not suitable for complex reasoning video tasks, e.g., referring video object segmentation~\cite{wu2022language}.

\section{Correlation between instance movement and stochastic process}
\label{sup_sec:brow}

Brownian motion and Brownian bridge are two classical stochastic processes. 
In Sec.~{``Introduction"}, we gradually analyze the instance movement via Brownian motion and Brownian bridge.
In this section, we first describe the definition of Brownian motion and Brownian bridge and elaborate on why they can model the instance movement.

\bmhead{Brownian Motion.}
Brownian motion, also called Winner process, is a continuous-time Gaussian stochastic process $W(t)$ which has stationary mean and linearly increasing variance over time. It can be formulated as:
\begin{align}
W(t)&\sim\mathcal{N}(0, t),              \label{sup_eq:brow_motion_density}\\
W(t) - W(s)&\upmodels W(r), 0 \leq r \leq s, \label{sup_eq:brow_motion_incre} \\
W(t) - W(s)&\sim\mathcal{N}(0, t-s),
\label{sup_eq:brow_motion_incre_density}
\end{align}
where Eq.~\ref{sup_eq:brow_motion_density} denotes the density function of Winner process at time $t$, Eq.~\ref{sup_eq:brow_motion_incre} denotes the increment of Winner process is independent~($\upmodels$) to itself, 
Eq.~\ref{sup_eq:brow_motion_incre_density} denotes the increment of Winner process follows a Gaussian distribution with a mean of $0$ and a variance of $t-s$.
Eq.~\ref{sup_eq:brow_motion_incre_density} also means the increment of Winner process is tiny in the early time.
As mentioned in Sec.~\textcolor{red}{1}, instance features of neighbor frames share highly similarity, which matches the property described in Eq.~\ref{sup_eq:brow_motion_incre_density}.
From this point, we initially model the instance movement as a Brownian motion.

\bmhead{Brownian Bridge.}
Brownian bridge~\cite{revuz2013continuous} is a continuous-time goal-conditioned Gaussian stochastic process $B(t)$ whose mean and variance are conditioned by start state $z_0$ at $t = 0$ and end state $z_T$ at $t = T$. It can be formulated as:
\begin{equation}
B(t)\sim\mathcal{N}((1 - \frac{t}{T})z_0 + \frac{t}{T}z_T, \frac{t(T-t)}{T}),
\label{sup_eq:brow_bri_density}
\end{equation}
where $z_t, t\in[0, T]$ is the middle state of the bridge. 
Considering that the instance movement has strong causal dependency between the middle state and start or end state, which means modeling instance movement as a Brownian motion disobeys Eq.~\ref{sup_eq:brow_motion_incre}. Then we find that the causal dependency accords with the goal-conditioned property of Brownian bridge so the instance Brownian motion is further modeled as an instance Brownian bridge.

\section{Brownian Bridge Distance Metric.}
\label{sup_sec:brow_bri_dist}

In this section, we elaborate on how does the distance metric is designed. Inspired by~\cite{wang2022language,zhang2023modeling}, we derive a contrastive objective to constrain instance features to mold instance features to follow Brownian bridge. The key of the contrastive objective is the definition of contrastive distance metric. In the Sec.~\textcolor{red}{4.3}, the distance metric is formulated as:
\begin{equation}
d(\mathbfcal{E}^{i}_{s},\mathbfcal{E}^{i}_{t},\mathbfcal{E}^{i}_{e}) = -\frac{1}{2\sigma^{2}}\left\Vert\mathbfcal{E}^{i}_t - (1 - \beta)\mathbfcal{E}^{i}_s - \beta\mathbfcal{E}^{i}_e\right\Vert^{2}_{2},
\end{equation}
where $\sigma^{2}$ is the variance in Eq.~\ref{sup_eq:brow_bri_density}: $\frac{(t - s)(e - t)}{(e - s)}$, $\beta$ is equal to $\frac{t-s}{e-s}$ and $i$ is the instance index at batch axis, i.e. $i\in [0, B\cdot N-1]$. 
If we denote instance embedding~$\mathbfcal{E}$ as instance state~$z$ and respectively set $s$ \& $e$ to $0$ \& $T$, the distance metrics can be further formulated as:
\begin{equation}
d(z_{0},z_{t},z_{T}) = -\frac{1}{2\sigma^{2}}\big\|
{z_t} - 
\underbrace{[(1 - \frac{t}{T})z_0 + \frac{t}{T}z_T]}_{\text{mean in Eq.~\ref{sup_eq:brow_bri_density}}} 
\big\|^{2}_{2}.
\label{sup_eq:brow_bri_dist}
\end{equation}
According to Eq.~\ref{sup_eq:brow_bri_dist}, we can find that the distance metric measures the distance between middle state~$z_t$ and mean value of Brownian bridge~$(1-\frac{t}{T})z_0 + \frac{t}{T}z_T$ at time $t$. When we maximize the distance metric, the middle state is moved to approach the center of Brownian bridge to follow the distribution of Brownian bridge.

\section{Dataset Analysis}
\label{sup_sec:dataset}

\subsection{Dataset Introduction}

Our method mainly utilizes five video instance segmentation datasets: YouTube-VIS 2019~\cite{yang2019video}, YouTube-VIS 2021~\cite{yang2019video}, OVIS~\cite{qi2022occluded}, BURST~\cite{athar2023burst}, and LV-VIS~\cite{wang2023towards}.

\bmhead{Youtube-VIS} is the most fundamental video instance segmentation benchmark, which contains 40 object classes. Youtube-VIS 2019 contains 2238/302/343 videos in training/validation/testing split. Youtube-VIS 2021 contains 2985/421/453 videos in training/validation/testing split and its annotation quality is higher than Youtube-VIS 2019.

\bmhead{OVIS} is a special video instance segmentation dataset in occluded scenes, which contains 25 object categories and has 607/140/154 videos in training/validation/testing split.

\bmhead{BURST} is a large-vocabulary video instance segmentation collected by extending TAO~\cite{dave2020tao} and involves 482 categories with 78 common categories from COCO~\cite{lin2014microsoft} and 404 uncommon categories.
It comprises 993/1421 in validation/testing split.

\bmhead{LV-VIS} is also a large-vocabulary video instance segmentation dataset, containing diverse object categories~(1196) and sufficient object mask annotations~(544k). Its training/validation split has 3076/837 videos.

\begin{table}[t]
\centering                         
\renewcommand{\arraystretch}{1.4}  
\setlength{\tabcolsep}{1.4mm}      
\footnotesize                      
\begin{tabular}{x{37}|x{30}x{30}y{30}y{30}y{30}}
\noalign{\hrule height 1.5pt}
Name     & YTVIS-19 & YTVIS-21 & BURST & OVIS  & LV-VIS \\
\hline
YTVIS-19 & 1.00     & 0.811    & 0.528 & 0.777 & 0.523  \\
LV-VIS   & 0.523    & 0.523    & 0.526 & 0.518 & 1.00   \\
\noalign{\hrule height 1.5pt}
\end{tabular}
\caption{The class similarity between training datasets~(YTVIS-19, LV-VIS) and evaluation datasets. Measured by Hausdorff distance and cosine similarity based on CLIP ViT-B/16 text encoder.}
\label{tab:cross_dataset_sim}
\end{table}

\subsection{Cross-Dataset Similarity}

Since we mainly adopt the ``cross-dataset' evaluation protocol to evaluate the open-vocabulary ability of methods, we hereby give a label-set similarity analysis between training and evaluation datasets to better understand our evaluation protocol.
Specifically, we adopt Hausdoff distance and cosine similarity based on text embeddings from CLIP ViT-B/16 text encoder to measure the cross-dataset similarity.
The results are summarized in Tab.~\ref{tab:cross_dataset_sim}.
When we choose YTVIS-19 as training dataset, YTVIS-21 and OVIS have a high-similarity score so they are more suitable for evaluating in-domain open-vocabulary ability~(i.e., invariant class set with variant visual scene). BURST and LV-VIS have a low-similarity score so they are mainly used for evaluating cross-domain open-vocabulary ability~(i.e., variant class set with variant visual scene).
When we choose LV-VIS as training dataset, YTVIS-19, YTVIS-21, OVIS, and BURST have a low-similarity score so all of them are used for evaluating cross-domain open-vocabulary ability.

\section{Extra Ablation}
\label{sup_sec:abla}

In this part, we mainly verify the effectiveness of our core ideas and design choices. Because the open-vocabulary ability is our main concentration, we conduct diagnostic experiments by training on LV-VIS and evaluating on BURST. Note that the BURST categories can be divided into three folds: ``all", ``common", and ``uncommon", we respectively denote the performance of these three parts as mAP$_{a}$, mAP$_{c}$, and mAP$_{u}$. mAP$_{a}$ reflects the overall open-vocabulary ability. mAP$_{c}$ reflects the vocabulary fitting ability and mAP$_{u}$ mainly exhibits the vocabulary transfer ability.

\begin{table}[h]
\centering                         
\renewcommand{\arraystretch}{1.4}  
\setlength{\tabcolsep}{1.5mm}      
\footnotesize                      
\begin{tabular}{x{33}x{33}|y{42}y{42}y{42}}
\noalign{\hrule height 1.5pt}
{Inter-frame} & {Intra-frame} & \multicolumn{1}{c}{{mAP$_{a}$}} & \multicolumn{1}{c}{{mAP$_{c}$}} & \multicolumn{1}{c}{{mAP$_{u}$}} \\
\hline
\cmark      &             & 6.21 & 9.56 & 5.37 \\
            & \cmark      & 5.92~\decrease{0.29} & 9.60~\increase{0.04} & 5.00~\decrease{0.37} \\
\rowcolor{aliceblue}
\cmark      & \cmark      & \textbf{6.53~\increase{0.32}} & \textbf{9.49~\decrease{0.07}} & \textbf{5.78~\increase{0.41}}\\
\noalign{\hrule height 1.5pt}
\end{tabular}
\caption{\textbf{Ablation Study} of TIR sub-modules, i.e., Inter-frame module and Intra-frame module. We follow the evaluation strategy of Tab.~\textcolor{red}{2}. Models are trained on LV-VIS dataset and are evaluated on BURST dataset. mAP$_{a}$, mAP$_{c}$, and mAP$_{u}$ denote mAP value on ``all", ``common", and ``uncommon" part of BURST val split.}
\label{tab:tir_abl}
\end{table}

\bmhead{Ablation of TIR}
Since the Inter-frame module and Intra-frame module serve different roles in TIR, we ablate the Inter-frame module and Intra-frame module from TIR in Tab.~\ref{tab:tir_abl} to verify the design choices of TIR. Singly adopting the Inter-frame and Intra-frame modules, their performance are promising~(6.21 mAP$_{a}$ and 5.92 mAP$_{a}$). Associating these modules provides a further improvement~(6.53 mAP$_{a}$), verifying the necessity of these two module designs. 

\begin{table}[h]
\centering                         
\renewcommand{\arraystretch}{1.4}  
\setlength{\tabcolsep}{1.5mm}      
\footnotesize                      
\begin{tabular}{x{20}x{20}x{20}|y{42}y{42}y{42}}
\noalign{\hrule height 1.5pt}
$\mathbfcal{L}_{htm}$ Eq.~\textcolor{red}{6} & $\mathbfcal{L}_{bc}$ Eq.~\textcolor{red}{9} & $\mathbfcal{L}_{btc}$ Eq.~\textcolor{red}{10} & \multicolumn{1}{c}{{mAP$_{a}$}} & \multicolumn{1}{c}{{mAP$_{c}$}} & \multicolumn{1}{c}{{mAP$_{u}$}} \\
\hline
       & \cmark & \cmark & 7.20                  & 9.72                 & 6.57 \\
\cmark &        & \cmark & 6.67~\decrease{0.53}  & 9.90~\increase{0.18} & 5.86~\decrease{0.71} \\
\rowcolor{aliceblue}
\cmark & \cmark & \cmark & \textbf{7.43~\increase{0.23}} & \textbf{9.53~\decrease{0.19}} & \textbf{6.91~\increase{0.34}}\\
\noalign{\hrule height 1.5pt}
\end{tabular}
\caption{\textbf{Ablation Study} of BTA sub-objectives, i.e., Head-Tail Matching~($\mathbfcal{L}_{htm}$, Eq.~\textcolor{red}{6}), Bridge Contrastive~($\mathbfcal{L}_{bc}$, Eq.~\textcolor{red}{9}), and Bridge-Text Contrastive~($\mathbfcal{L}_{btc}$, Eq.~\textcolor{red}{10}). The experiment settings are equivalent to Tab.~\ref{tab:tir_abl}. }
\label{tab:bta_abl}
\end{table}

\bmhead{Ablation of BTA}
BTA serves as a comprehensive training objective~(Eq.\textcolor{red}{11}), containing three sub-objectives, i.e., Head-Tail Matching~($\mathbfcal{L}_{htm}$), Bridge Contrastive~($\mathbfcal{L}_{bc}$), and Bridge-Text Contrastive~($\mathbfcal{L}_{btc}$). Because $\mathbfcal{L}_{btc}$ is necessary for classification, we mainly ablate $\mathbfcal{L}_{htm}$ and $\mathbfcal{L}_{bc}$ from BTA in Tab.~\ref{tab:bta_abl}. We can find that $\mathbfcal{L}_{bc}$ has already achieved promising results~(7.20 mAP$_{a}$) and the extra introduction of $\mathbfcal{L}_{htm}$ further improves its performance~(7.20 $\rightarrow$ 7.43 mAP$_{a}$), demonstrating the effectiveness of these designs.
In addition, the single introduction of $L_{htm}$ can improve baseline w/ TIR slightly~(6.53 $\rightarrow$ 6.67 mAP$_{a}$).

\begin{figure}[h]
\centering
\includegraphics[width=1.0\linewidth]{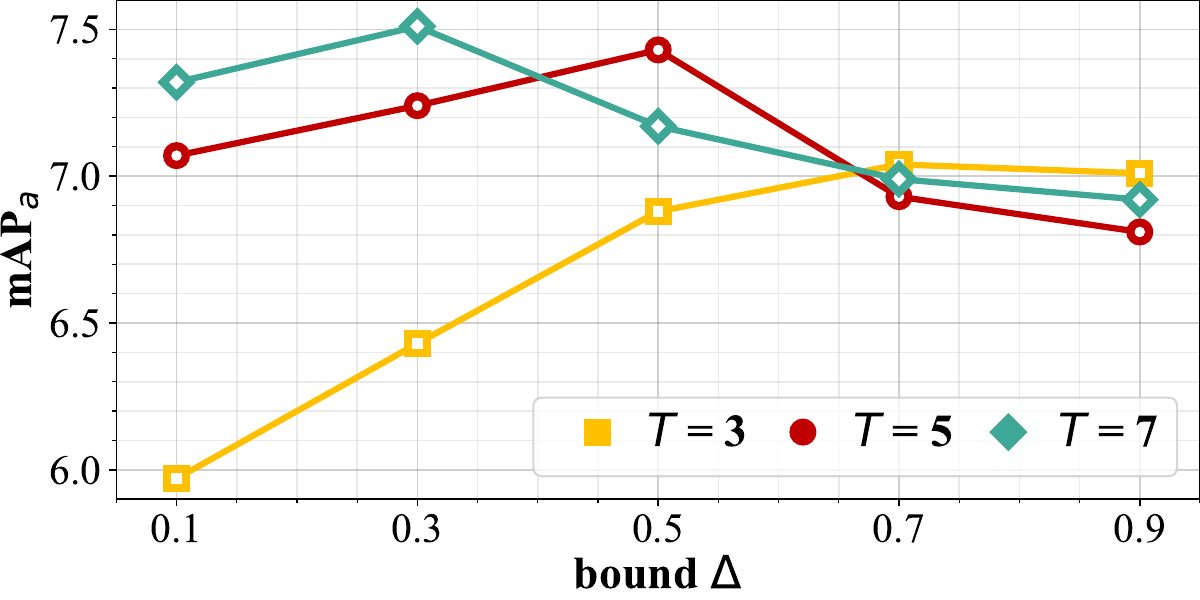} 
\caption{\textbf{Parameter Sensitivity Analysis} of hyperparameters: matching margin $\Delta$ and frame sampling number $T$. The default value of $\Delta$ and $T$ are set to 0.5 and 5, respectively. 
}
\label{fig:param_sensi}
\end{figure}

\bmhead{Parameter Sensitivity}
Learning of the bridge modeling ability is controlled by two hyperparameters~($\Delta$, $T$). $\Delta$ controls the bridge width in feature space. $T$ controls the semantic change across the video. We hope that the bridge width can exactly reflect the semantic change so the values of $\Delta$ and $T$ need to match with each other. We mainly evaluate our model when $\Delta\in[0.1, 0.9]$ with $T=3,5,7$. As shown in Fig.~\ref{fig:param_sensi}, we can find that smaller $\Delta$ requires larger $T$, which also shows better ultimate performance~(e.g., 7.01 mAP$_{a}$ for $\Delta=0.3,T=3$, 7.43 mAP$_{a}$ for $\Delta=0.5,T=5$, 7.54 mAP$_{a}$ for $\Delta=0.7,T=7$). Taking performance and efficiency into account, we choose $\Delta=0.5,T=5$ as our default parameter value.

\end{document}